\documentclass[final]{article}
    \makeatletter
    \def\ps@pprintTitle{%
    \let\@oddhead\@empty
    \let\@evenhead\@empty
    \def\@oddfoot{}%
    \let\@evenfoot\@oddfoot}
    \makeatother
    \usepackage{lineno}
    \usepackage{mathrsfs}
    \usepackage[breaklinks=true,pdftex]{hyperref}
    \modulolinenumbers[1]
    \usepackage{amsmath,bm}
    \usepackage{svg}
    \usepackage{comment}
    \usepackage{multirow}
    \usepackage[a4paper,
            bindingoffset=0.1in,
            left=1.2cm,
            right=1.2cm,
            top=1.5cm,
            bottom=1.5cm]{geometry}
    \usepackage{ulem}
    \usepackage{booktabs}
    \usepackage{dirtytalk}
    \usepackage{lineno}
    \usepackage{caption}
    \usepackage{amssymb}
    \usepackage{amsmath}
    \usepackage{savesym}
    \usepackage{algorithmic}
    \usepackage{algorithm}
    \savesymbol{AND}
    \usepackage{adjustbox}
    \usepackage{tabularx} 
    \usepackage{graphicx} 
    \usepackage{pdflscape}
    \usepackage{rotating}
    \usepackage{wrapfig}
    \usepackage{amssymb}
    \usepackage{soul,color}
    \usepackage{subcaption}
    \usepackage{lineno}
    \usepackage{xcolor}
    \usepackage{url}
    \soulregister\cite7
    \soulregister\ref7
    \soulregister\pageref7

    \usepackage{todonotes}

    \DeclareRobustCommand{\uvec}[1]{{%
    \ifcsname uvec#1\endcsname
    \csname uvec#1\endcsname
    \else
    \bm{\hat{\mathbf{#1}}}%
    \fi}}
    \mathchardef\breakingcomma\mathcode`\,
    {\catcode`,=\active
      \gdef,{\breakingcomma\discretionary{}{}{}}
    }
    \newcommand{\mathlist}[1]{$\mathcode`\,=\string"8000 #1$}
    \begin{document}
    \begin{center}
    {\Large \textbf{Physics-Informed Geometric Operators to Support Surrogate, Dimension Reduction and Generative Models for Engineering Design}}\\
    \vspace{0.6cm}
    {\small Shahroz Khan$^{1*}$\let\thefootnote\relax\footnote{$^*$Corresponding author. E-mail address: shahroz.khan@bartechnologies.uk (S. Khan)}},
    {\small Zahid Masood$^{2}$},
    {\small Muhammad Usama$^{3}$},
    {\small Konstantinos Kostas$^2$},
    {\small Panagiotis Kaklis$^{3,4}$}, and
    {\small Wei (Wayne) Chen$^{5}$}
    \\\vspace{0.2cm}
    {\small $^1$BAR Technologies, Portsmouth, UK}\\
    {\small $^2$Department of Mechanical and Aerospace Engineering, Nazarbayev University, Astana, Kazakhstan}\\
    {\small $^{3}$Department of Naval Architecture, Ocean and Marine Engineering, University of Strathclyde, Glasgow, UK}\\
    {\small $^4$Foundation for Research \& Technology Hellas (FORTH), Institute of Applied \& Computational Mathematics (IACM), \\ Division: Numerical Analysis \& Computational Science, Group: Data Science, Heraklion, Crete, Greece}\\
    {\small $^{5}$J. Mike Walker ’66 Department of Mechanical Engineering, Texas A\&M University, Texas, USA}\\
    \end{center}
    
    \section*{\centering Abstract}
    In this work, we propose a set of physics-informed geometric operators (GOs) to enrich the geometric data provided for training surrogate/discriminative models, dimension reduction, and generative models, typically employed for performance prediction, dimension reduction, and creating data-driven parameterisations, respectively. However, as both the input and output streams of these models consist of low-level shape representations, they often fail to capture shape characteristics essential for performance analyses. Therefore, the proposed GOs exploit the differential and integral properties of shapes—accessed through Fourier descriptors, curvature integrals, geometric moments, and their invariants—to infuse high-level intrinsic geometric information and physics into the feature vector used for training, even when employing simple model architectures or low-level parametric descriptions. We showed that for surrogate modelling, along with the inclusion of the notion of physics, GOs enact regularisation to reduce over-fitting and enhance generalisation to new, unseen designs. Furthermore, through extensive experimentation, we demonstrate that for dimension reduction and generative models, incorporating the proposed GOs enriches the training data with compact global and local geometric features. This significantly enhances the quality of the resulting latent space, thereby facilitating the generation of valid and diverse designs. Lastly, we also show that GOs can enable learning parametric sensitivities to a great extent. Consequently, these enhancements accelerate the convergence rate of shape optimisers towards optimal solutions.
    \vspace{0.2cm}\\
    \textit{Keywords:} Surrogate/Discriminative Models,  Dimension Reduction, Generative Models, Sensitivity Analyses, Parametric Design, Shape Optimisation.

    \section{Introduction}\label{INTRO}
    The computational costs in simulation-driven optimisation are predominantly attributed to two factors: the high dimensionality of the design space and high-fidelity solvers aimed at replicating physical phenomena. High-fidelity solvers, regardless of the design space's dimensionality, add a computational burden to the optimisation process \cite{shukla2024deep}. This challenge is further amplified in high-dimensional design spaces due to the increased number of time-consuming simulations that are generally required. Such issues related to the high-dimensionality are typically categorised under the \say{curse of dimensionality,} which exponentially increases the complexity of the optimisation problem with increasing numbers of parameters \cite{poole2017high}.
     
    In response, data-driven approaches, such as machine/deep learning (ML) models, have become central in efforts to mitigate this complexity. Although high-performance computing has significantly contributed to accelerating the involved physical simulations, a significant part of the increased computational power is still consumed in physical solvers which have simultaneously advanced in their complexity and fidelity in replicating real-world physics. Consequently, the need for data-driven models, which have been treated as black boxes to a large extent by the engineering and science community, remains critical. Recently, however, this community started advocating for models that are intrinsically connected with the physical phenomena they aim to replicate. An example of this trend is the emergence of physics-informed models \cite{karniadakis2021physics}, which have arguably led to the development of scientific ML.
    
    Scientific ML models are used to either reduce computational load, by creating a low-fidelity substitute (surrogate models (SM)) for the original physical simulation, or decrease the dimensionality of the design space. The latter is accomplished either by selecting a subset of parameters/features, which contribute significantly to the performance metric~\cite{nikishova2020sensitivity}, or extracting latent features for a subspace that replicates the original high-dimensional design space \cite{diez2015design}. Typically, the dimensionality of subspaces is significantly lower than that of the original space, thereby requiring fewer resource-intensive simulations during optimisation to potentially reach a global optimum. 

    SMs are typically built using discriminative ML models such as random forests, gradient boosting machines, support vector machines with Gaussian kernels, Gaussian processes, neural Networks (NN), and their convolutional (CNN) or graph (GNN) variants \cite{goodfellow2016deep}. It is noteworthy that the term \say{Discriminative models} has been historically associated with classification tasks, but such approaches are also applicable in surrogate modelling settings \cite{weisz2023toward,gordon2020combining}. These models are typically supervised and require a dataset $D=\lbrace (x_n, y_n) \rbrace_{n=1}^N$ consisting of input-output pairs to model the conditional distribution $\mathrm{p}(y|x_n)$ of the outputs given the inputs. This process involves optimising a set of parameters, $\Theta$, during training. 
    
    While ML-based models are preferred for complex physical systems, traditional methods such as linear/nonlinear regression, radial basis functions, splines, and piece-wise interpolation are becoming less favoured \cite{ryan1997modern}. These can be categorised as non-ML-based methods because they do not involve iterative learning from the data and are typically designed based on predefined mathematical formulations. They are simpler and do not adapt over time with new data inputs, marking a significant distinction from ML-based methods.

    On the other hand,  dimension reduction models (DRMs), used for feature extraction-based subspaces, are typically unsupervised in nature \cite{diez2015design}. They require only the design data $D=\{x_n\}_{n=1}^N$ and employ ML-based models such as auto-encoders, deep belief networks, and locally linear embedding, as well as linear discriminant analysis \cite{goodfellow2016deep} for processing them. Other widely used DRMs, such as singular value decomposition (akin to proper orthogonal decomposition or principal component analysis), factor analysis, and shape-signature vectors \cite{KHAN2022103327,10.3390/jmse11101851}, are also non-ML-based, as they do not involve iterative learning, akin to non-ML SMs. In contrast, feature selection-based DRMs are supervised and focus on learning the impact/significance of each feature with respect to the quantity of interest, e.g., the design's physical performance. Once the significance of each feature is identified, the dimensionality of the design space is reduced by eliminating or fixing the values of features with low impact on the quantity of interest.
    
    Despite the proven efficiency of ML approaches in expediting the design pipeline, particularly in optimisation settings, their ability to generate innovative solutions has been limited. This limitation stems from the fact that they primarily focuses on predicting performance criteria of designs from rather limited and narrow design spaces, or on learning latent features for subspace construction~\cite{KHAN2022103327}. The advent of modern generative models (GM)~\cite{tomczak2022deep} and their application in engineering have significantly altered the landscape by offering the possibility of rich and efficient design spaces. The generative design spaces enable the creation of innovative shapes, in addition to performance prediction. Unlike SMs, GMs possess the capability to generate valid, unseen and novel data, i.e., designs in our context. They achieve this by capturing the joint probability distribution $\mathrm{p}(X, Y)$, or just $\mathrm{p}(X)$ in scenarios where labels are absent~\cite{gordon2020combining}. The latter scenario is often preferable in engineering applications since evaluating performance labels can be computationally intensive. These models empower designers and engineers to develop data-driven design parameterisations, thereby facilitating a rich design space for physics-informed design exploration. This approach enables the generation of optimal alternatives beyond conventional imagination, integrating high-level design objectives and paving the way for not only innovative but also physically high-quality solutions~\cite{khan2023shiphullgan,chen2021padgan}.
    
\begin{figure*}[hbt!]
    \centering
    \includegraphics[width=0.9\textwidth]{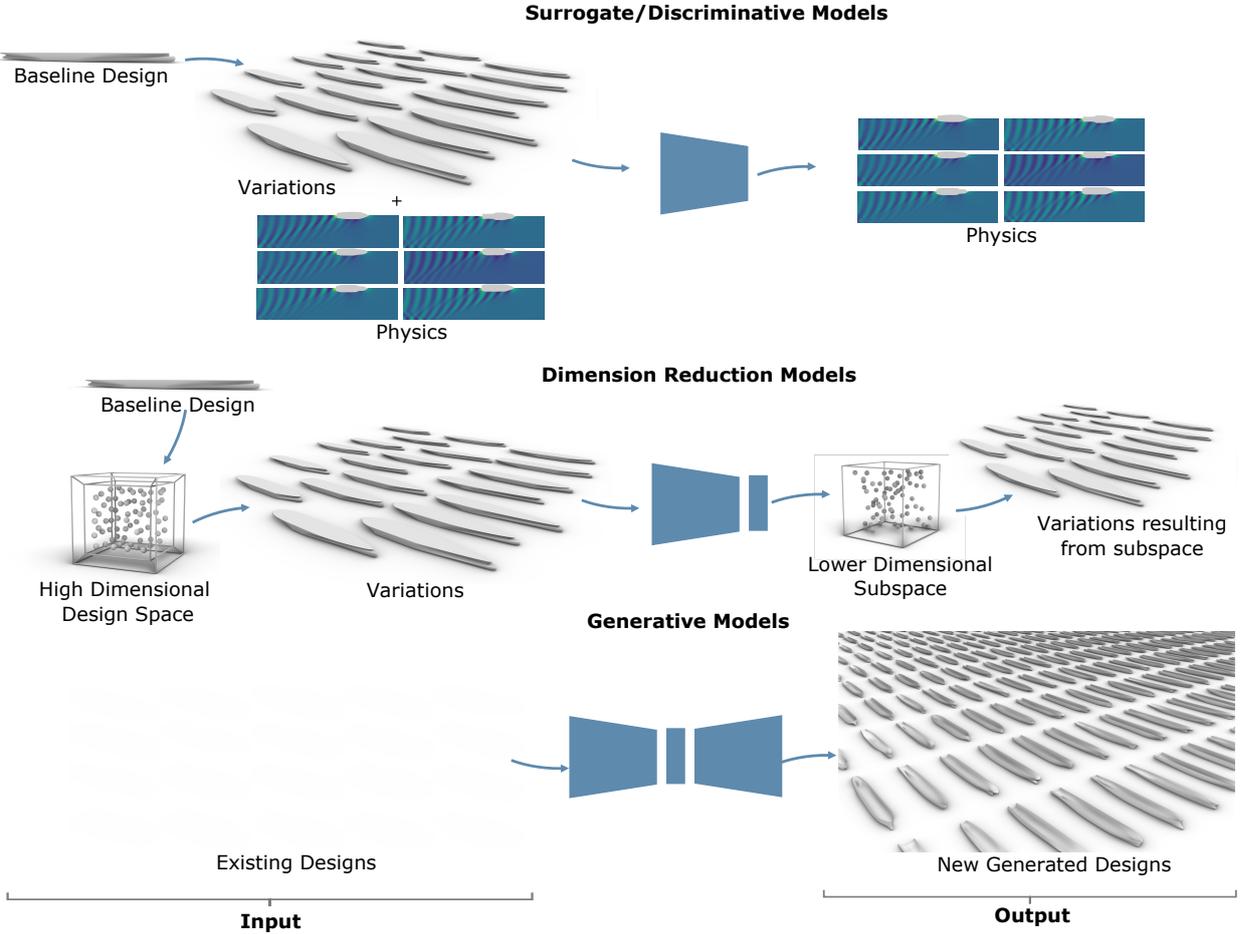}
    \caption{Illustrative layout of typical surrogate/discriminative models (SMs), dimension reduction models (DRMs), and generative models (GMs). SMs are supervised models that require both design and performance data for training, which are later used to predict the performance of new designs. DRMs, on the other hand, are unsupervised; they only need the original designs to construct a lower-dimensional latent space. GMs are also unsupervised models where both input and output are designs. They are trained using a set of existing designs, adopting a probabilistic approach that allows the creation of new and unseen designs adhering to the learned patterns.}\label{training_example}
    \end{figure*}

    Applications of DRMs and SMs for expedited simulation-driven optimisation are widely documented in pertinent literature~\cite{iliadis2023engineering}. However, the integration of GMs in engineering design has only recently begun. This shift can be largely attributed to the introduction of variational autoencoders, followed by advancements in generative adversarial networks, diffusion models, and transformers~\cite{regenwetter2022deep}. While there are other traditional ML (like latent Dirichlet allocation and auto-regressive models) and non-ML (such as Hidden Markov and Agent-based models) generative models \cite{rubinstein1997discriminative}, their footprint in engineering design is almost negligible. It is safe to say that these models have remained relatively unknown to the design community, which has recently shown an interest in generative AI. In contrast, non-ML grammar-based generative methods~\cite{granadeiro2013general} have seen extensive use in design. These models employ a set of rules, the \textit{grammar}, to generate valid and diverse designs. A list of widely used SM, DRM and GM approaches is given in Table \ref{table_1}, and a graphical layout of their basic training process is shown in Figure~\ref{training_example}.
    
    \begin{table}[htb!]
    \small
    \centering
    \caption{List of some prominent SM, DRM and GM approaches.}
    \begin{tabularx}{\linewidth}{lXX}
     \toprule
     & ML-based models & Non ML-based models \\
     \toprule
    Discriminative/Surrogate & random forest, gradient boosting machines, support vector machines, gaussian kernels, neural networks, convolutional NN and graph NN &  linear/nonlinear, kriging, logistic , polynomial, and ridge/lasso regression \\\\
    Generative & variational auto-encoders, 
    generative adversarial networks, diffusion models, transformers, latent Dirichlet allocation and autoregressive models & hidden Markov, agent- and grammar-based models \\\\
    & \multicolumn{2}{c}{\textbf{Subspace learning (feature extraction)}}\\
    Dimension Reduction &  auto-encoders, deep belief networks, locally linear embeddings, linear discriminant analysis &  singular value decomposition (proper orthogonal decomposition, principal component analysis), factor analysis, and shape signature vectors\\\\
     & \multicolumn{2}{c}{\textbf{Feature selection}}\\
    & wrapper (forward selection, backward elimination, recursive feature elimination), embedded approaches & sensitivity (variance, derivative, Morris, One-At-A-Time), filtering (Pearson correlation, LDA)\\
     \bottomrule
    \end{tabularx}
    \label{table_1}
    \end{table}
    
    \subsection{Challenges in training surrogate, generative and dimension reduction models}\label{drawback}
    
    Despite their inherent capacity in modelling complex relations, these models require a well-compiled dataset to deliver meaningful responses~\cite{sindhu2020survey,regenwetter2022deep}. Traditionally, such models are designed to mainly handle datasets with one-dimensional (1D) or two-dimensional (2D) records representing financial time series, words and word segments from natural languages, and images as a whole or in parts. The intrinsic data representation in these domains (numeric, text and pixel information), is well-studied and established to a large extent. However, representing three-dimensional (3D) free-form shapes poses a significant challenge for SMs, GMs and DRMs. Specifically, transforming/encoding these 3D shapes into the commonly used 1D or 2D records varies greatly with equally varying outcomes, and more importantly, such transformations might result in inadequate representations. Obviously, the ultimate aim is to represent such shapes in a computationally efficient way which also captures the structure needed to accurately estimate the relevant performance metrics.
    
    In engineering design problems, SMs typically process input data in the form of a finite list of parametric values, describing the shape, followed by a smaller set of values or a distributed physical quantity corresponding to the quantity of interest for each design. While such vector descriptors efficiently define the shape, they commonly fail to capture intrinsic geometric structures and features. This omission may lead to a loss of important spatial and geometric information, such as topology, curvature, depth, or surface-smoothness, which can be essential for accurate performance predictions. Therefore, in this case, the resulting models may excel at reconstructing existing designs or close variants, but typically struggle with producing truly innovative shapes that do not appear in the training data. 
    
    To address this issue, convolutional- and graph-based SM models, employing a variety of geometric representations like meshes, signed distance functions, voxels, or point clouds, have been used in literature. These approaches can capture the missing pieces of geometrical information of the design to some extent, but they commonly suffer from the increased complexity at the model level and the shape preprocessing which can be labour-intensive, making reliable training challenging. Additionally, interpreting the resulting models' behaviour becomes equally challenging, depriving engineers' of any intuitive notion of why specific shape descriptors perform better or worse. This lack of interpretability can make design optimisation challenging and essentially restricts engineers in producing small incremental improvements in the vicinity of existing designs. Finally, when such models are employed, costly conversions between model-specific and analysis-suitable representations are needed during performance evaluation, adding further complexity to the overall optimisation pipeline. Nevertheless, training SMs with design vectors, comprising parameter values, ensures smooth integration with optimisers in the optimisation loop, but as mentioned earlier, these representations are often inadequate approximations leading to potential losses in model accuracy.
    
    When GM and DRM are considered, there is a clear alignment in the sense that both extract latent features from the given design datasets; however, these features are of different types. Specifically, GMs extract features for modelling the underlying data distribution, thereby enabling them to generate novel designs, whereas DRMs focus solely on extracting latent features for reducing dimensionality without explicitly modelling the underlying data distribution.  However, due to the lack of correlation among design parameters, training with the aforementioned parameters can yield perfunctory latent subspaces. In these subspaces, the features merely form a new orientation of the design space without capturing any associated geometric features. Consequently, these models are typically trained directly with design representations, such as surface discretisations.
    
    Since both the input and output streams of these models are limited to low-level 3D shape representations, they often fail to capture the structural and shape characteristics crucial for analysing performance. Consequently, lack of surface smoothness and large number of invalid designs, e.g., non-watertight or self-intersecting designs, are common problems in the generated designs. This is particularly critical in engineering analysis where maintaining surface smoothness and validity is vital, as even minor local variations in surface quality can significantly impact performance. In addition, due to their unsupervised nature, these models fail to incorporate any notion of physics. 
    
    In common computer graphics applications, surface smoothness is rarely critical since most tasks only require the generation of geometric objects that will be perceived as realistic by human agents, and subsequent refinements can be applied if and when detailed analysis is needed. On the contrary, when engineering analysis with free-form shapes is considered, surface smoothness (commonly tangent-plane continuity) is essential for performance evaluation. For example, in the case of ship hulls, wings or turbine blades, surface smoothness does not serve aesthetic criteria but ensures fluid flow characteristics that are essential for the design performance. Although smoothness could be potentially achieved by an increased resolution and/or refinements, this would lead to an explosion of model's complexity and memory requirements.
    
    In summary, the inability of the abovementioned approaches to extract appropriate geometric or physics-associated features is not inherently due to the models themselves but largely arises from the shape descriptors used for training, which are often low-level shape discretisations. Therefore, extracting appropriate latent features from such representations is extremely challenging, if not infeasible, and consequently underlines the need for richer representations which would embed the underlying shape’s structure, features, and physics.
    
    \subsection{Proposed approach}
    In this work, we propose the use of physics-informed geometric operators (GOs) to enrich the geometric data provided to the employed models. We claim that this addition enables the extraction of useful high-level shape characteristics, even when simple model architectures or low-level data representations, like design parameters or surface discretisation, are used. GOs leverage the shape's differential and integral properties, which are retrieved via Fourier descriptors, curvature integrals, geometric moments, and their invariants, to capture the varying characteristics of the shape related to its volume distribution, complexity of the bounding surface, and overall surface smoothness. These operators introduce high-level intrinsic geometric information and physics in the resulting shape descriptors. These operators capture both global and local shape features by explicitly encoding the relevant shape information which not only augments the training dataset with a compact geometric representation of free-form shapes but also embeds physical information. The latter is achieved through the selection of shape characteristics that are correlated with design's performance (wave-making resistance, lift, drag etc.), thereby making the model, physics-informed. However, studies of such approaches for ML models applied in engineering design are extremely limited. 
    
    Our motivation to use GOs stems from their application in shape descriptors used in both traditional and ML methods for classification, object recognition and retrieval tasks~\cite{reeves1988three}. For example, in ML, Joseph-Rivlin et al.~\cite{joseph2019momen} proposed a Moment-enriched model, Momen$^e$t, for point-cloud classification, which was later modified by Li et al.~\cite{li2020ggm} to produce a graph geometric moments convolution neural network, GGM-Net, which learns local geometric features from geometric moments representation of a local point set. Geometric moments can capture shape characteristics at different levels with low-/high-order moments capturing low-/high-order shape characteristics, respectively. This approach has been extensively used in estimating surface similarity since the coincidence of all moments between two surfaces, implies identical surfaces. As for differential characteristics, Ye et al.~\cite{ye2021curvature} proposed a generative model for 3D surfaces based on a shape representation including mean curvature values.

    Another well-known feature of the proposed GOs is their strong coupling with physics phenomena underlying the performance characteristics of functional free-form surfaces. This is rather obvious if we consider performance indices, such as lift and wave-making resistance, as the selected GOs supply important clues about the form and its distribution, as well as the validity of the design~\cite{gmdsa_r5, ssdr_r5,shen2017computational}. However, contrary to most performance indicators, their evaluation is substantially less expensive. Therefore, in this work, we use GOs to demonstrate that such augmented shape descriptors are beneficial both with respect to the resulting model's performance as well as the model's capacity to act as a physics-informed one. Especially, for GMs and DRMs, the incorporation of the proposed GOs significantly improves the resulting latent space’s compactness, thereby facilitating the generation of valid designs from latent design spaces. Consequently, this enhancement accelerates the convergence rate of shape optimisers towards optimal solutions.
    
    Our experiments further demonstrate that GO-augmented shape descriptors result in measurable improvements in modelling accuracy and enhancements in the model’s generalisation capability. Various finite combinations of GO-induced values can obviously capture different sets of underlying geometric information that are herein studied in their efficacy in acting as sufficient and compact signatures of the corresponding shape surfaces. 
    
    \section{Geometric operators}
    This section details the proposed method, including the basic assumptions and the mathematical formulation of GOs. It also covers the theoretical basis and the steps for its computational implementation, aiming to provide a clear understanding of the foundation of the approach and its application in practice.
    
    \subsection{Problem formulation}
    Let ${\cal G}$ be a geometric object representing a baseline design in an ambient space ${\cal A}\subseteq\mathbb{R}^3$. We also assume that $\mathcal{P}({\cal G})$ is a vector function in a finite space that provides a suitable representation, $\mathbf{x} = \mathcal{P}({\cal G})\in \mathbb{R}^{n_P}$, of ${\cal G}$ in $\mathcal{A}$. Here, $\mathcal{P}$ can be either a suitable continuous parameterisation of ${\cal G}$ or its geometric approximation in the form of a point discretisation, mesh, voxelisation, or similar. In addition, we consider three lumped operator vectors, $\mathcal{M}({\cal G}) \in \mathbb{R}^{n_\mathcal{M}}$, $\mathcal{K}({\cal G}) \in \mathbb{R}^{n_K}$ and ${\cal F}({\cal G}) \in \mathbb{R}^{n_F}$ representing geometric moments, curvature integrals, and Fourier descriptors of $\mathcal{G}$, respectively. Combining these additional vectors with   $\mathbf{x} = \mathcal{P}({\cal G})$ results in a well defined geometric operators (GOs),
    
    \begin{equation}\label{GO_1}
       \mathrm{GO}(\mathcal{G}) =\left(
       \mathcal{P}({\cal G}),~
       \mathcal{M}({\cal G}),~
       \mathcal{K}({\cal G}),~
       \mathcal{F}({\cal G})
       \right).
    \end{equation}
    
\begin{figure*}[hbt!]
    \centering
    \includegraphics[width=0.9\textwidth]{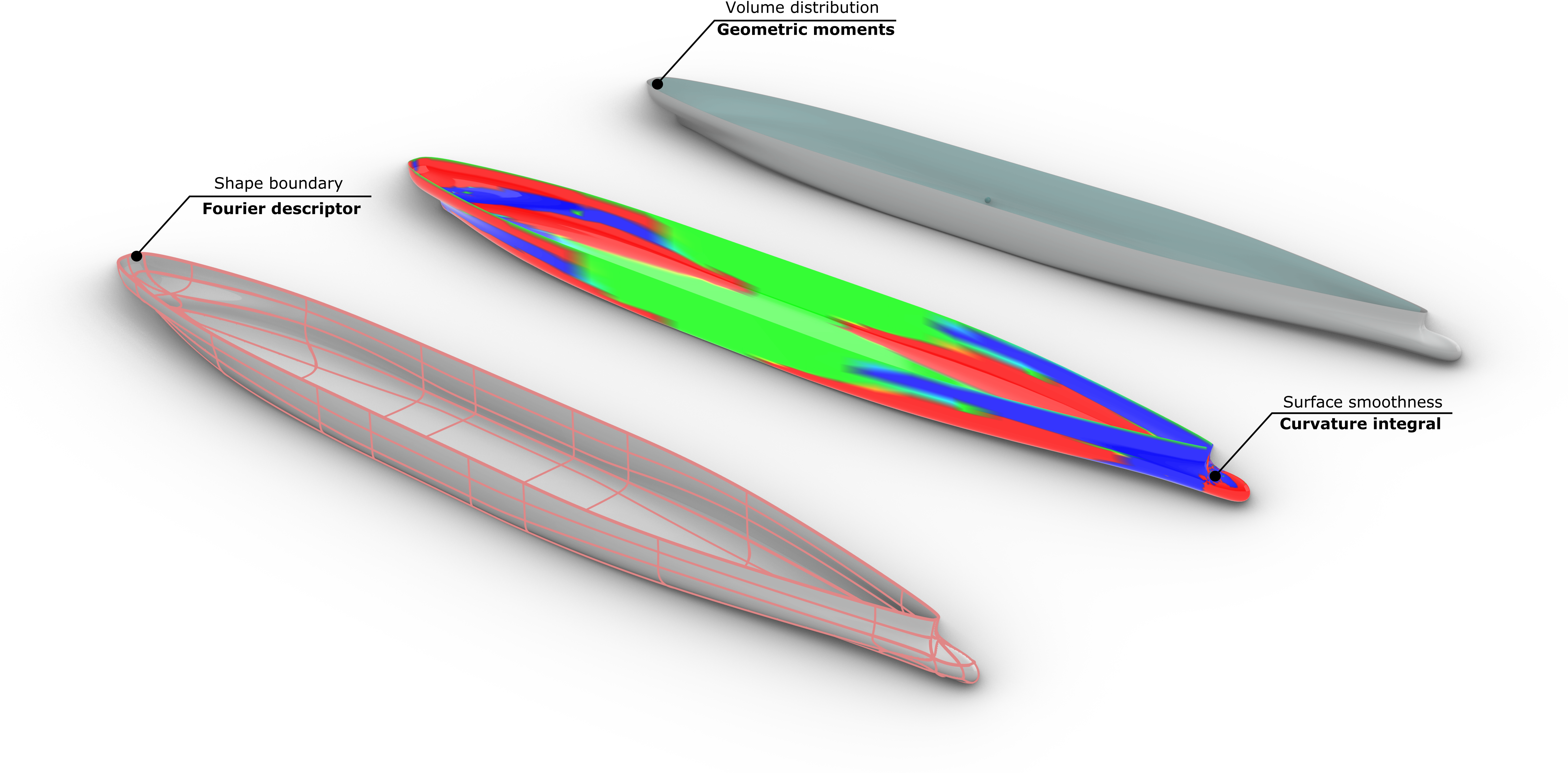}
        \caption{Illustration of various geometric characteristics captured with geometric moments-, curvature integral- and Fourier descriptor-based geometric operators.}
    \label{GO_rep}
    \end{figure*}

    \noindent  If the initial shape dataset contains $n$ designs, $\{$\mathlist{\mathbf{x}_1,\mathbf{x}_2,\mathbf{x}_3,\dots,\mathbf{x}_n}$\}$ with $\mathbf{x}_i=\mathcal{P}(\mathcal{G}_i)$, the resulting GO-based training dataset is transformed into $\mathcal{X}=\{$\mathlist{\mathrm{GO}_1,\mathrm{GO}_2,\mathrm{GO}_3,\dots,\mathrm{GO}_n}$\}$, with $\mathrm{GO}_i = \mathrm{GO}(\mathcal{G}_i)$. For all models, SM, GM or DRM, the components of $\mathrm{GO}(\mathcal{G})$ are the input features fed into their respective architectures, with each subset of components, $\mathcal{P}, \mathcal{M}, \mathcal{K}$ and $\mathcal{F}$, capturing a different set of shape characteristics; \textcolor{red}{ see Figure \ref{GO_rep} for illustration.} For example, $\mathcal{M}({\cal G})$ captures the shape volume and its distribution in space, while $\mathcal{F}({\cal G})$ and $\mathcal{K}({\cal G})$ are describing the shape's bounding surface characteristics, in relation to its complexity and smoothness, which will be discussed in more detail in the subsequent subsections. At the same time, and more importantly, these operators enact regularisation by ensuring that the trained models pay attention to the respective characteristics, thereby reducing over-fitting effects and facilitating better generalisation to new, unseen designs. 

    \subsection{Geometric moments}
    Geometric moments are well-known for their ability to capture shape characteristics at various levels. This translates to a wide range of comprehensive geometric features available for various shape processing tasks, including applications in probability and statistics~\cite{diaconis1987application}, object recognition~\cite{ssdr_r34}, rigid body transformations~\cite{gmdsa_r8}, image analysis~\cite{teh1988image} and computational tomography~\cite{milanfar1995reconstructing} among many others. Another important aspect, particularly valued in the engineering community, is their strong connection to physical quantities that commonly determine the design's performance characteristics. Geometric moments appear in the equations of various physical quantities used in material field modelling~\cite{ssdr_r29}, fluid flows around bodies~\cite{ssdr_r7}, and, more recently, in mesh-free finite element analysis (FEA)~\cite{ssdr_r5}, where moment-based shape representations promote interoperability between CAD representations and physics simulations. In summary, the inclusion of geometric moments in GOs is motivated by the following fundamental insights:
    \begin{enumerate}
    \item Shape's geometric moments capture its intrinsic geometric properties and act as a unifying medium between geometry and its physical evaluation~\cite{gmdsa_r5, ssdr_r5}.
    \item A large number of governing equations and quantities of interest in engineering analysis are related to these integral geometric properties and are commonly used to construct efficient computational methods~\cite{ssdr_r5, ssdr_r7}.
    \item Similar to their role in physics, geometric moments also act as a compact shape signature or descriptor, facilitating various shape processing tasks~\cite{gmdsa_r8, ssdr_r34}.
    \end{enumerate}
    
    \subsubsection{Computation of geometric moments}    
    In view of the above, let us start with the definition of an $s-$order, $s=p+q+r$, geometric moment $M^s=M^{p,q,r}$ of a geometric object $\mathcal{G}$ in $\mathcal{A}\subseteq\mathbb{R}^3$, given by
    \begin{equation}\label{eq:Mpqr}
        M^{p,q,r}(\mathcal{G}) = \iiint_{\cal G}x^p\,y^q\,z^r\,\text{d}x\,\text{d}y\,\text{d}z, \qquad \mathrm{with} \qquad p,q,r\in\lbrace 0,1,2,\dots\rbrace.
    \end{equation}
    Ideally, the selection of $s$ should result in an appropriate set of geometric moments capturing global as well as local features of $\mathcal{G}$. Drawing on the well-established theory of moments~\cite{shohat1950problem}, we do expect geometric moments, as defined in \eqref{eq:Mpqr}, to capture significant information about the underlying geometry and, under specific conditions, even achieve unique reconstruction~\cite{gmdsa_r58,gmdsa_r59}. Zero- and first-order moments, $M^{0,0,0}$ and $\lbrace M^{1,0,0}$, $M^{0,1,0}$, $M^{0,0,1} \rbrace$, are the most commonly used moments in computer graphics, CAD and engineering for computing a shape's volume, $\mathcal{V}=M^{0,0,0},$ and its centroid, $\mathbf{c}=(c_x,c_y,c_z)=\left(\frac{M^{1,0,0}}{M^{0,0,0}},\frac{M^{0,1,0}}{M^{0,0,0}},\frac{M^{0,0,1}}{M^{0,0,0}}\right)$. The rank-2 tensor of second order moments, also refereed to as the moment of inertia tensor, is commonly used in engineering analyses and applications. As one might expect, the more moments we use, the better we capture the shape's intrinsic features. Therefore, one may opt for the inclusion of all moments up to $s$-order moments, i.e.,
      \begin{equation}\label{mf}
        \bm{M}^s = \{M^{p,q,r}:\ p+q+r = s\},
    \end{equation}
    with $s$ being appropriately large to cover the shape characteristics of interest~\cite{gmdsa_r8}. Theoretically, $s$ ranges from $0$ to $\infty$, though there exist object classes for which $s$ is finite when, e.g., dealing with the class of the so-called quadrature domains in the complex plane~\cite{gmdsa_r58} or when approximating convex bodies using Legendre moments~\cite{gmdsa_r59}. In light of this, we define the moment-based operator of size $n_\mathcal{M}$, as the vector comprising all elements of vectors $\bm{M}^j$ for $j=0,1,2,...,s$,
    \begin{equation}\label{eq:SSV}
        \mathcal{M} = \lbrace\bm{M}^0, \bm{M}^1, \bm{M}^2, \dots, \bm{M}^s\rbrace,\quad\mathcal{M}\in\mathbb{R}^{n_\mathcal{M}},
    \end{equation}
where, for example $\bm{M}^0 = \lbrace M^{0,0,0}\rbrace$, $\bm{M}^1 = \lbrace M^{1,0,0}, M^{0,1,0}, M^{0,0,1}\rbrace$ and $\bm{M}^2=\{$\mathlist{M^{2,0,0}, M^{1,1,0}, M^{1,0,1}, M^{0,2,0}, M^{0,1,1}, M^{0,0,2}}$\}$. Therefore, since all $\bm{M}^j$ are disjoint,
    \begin{align}\label{eq:SSVcardinality}
        n_\mathcal{M}=\sum_{j=0}^{s}\text{Card}(\bm{M}^j)= \sum_{j=0}^{s}\frac{(j+1)(j+2)}{2}= \frac{s^3}{6}+{s}^2+\frac{11s}{6}+1.
    \end{align}
If, for example, central moments are considered, one may want to exclude $\bm{M}^1$ since in that case all of its components would be equal to 0. Hence, if $\bm{M}^1$ is excluded, i.e., $j=0,2,3,...,s$, then $n_\mathcal{M}$ becomes equal to $\left(\frac{s^3}{6}+{s}^2+\frac{11s}{6}-2\right)$.

A variety of methods is available in the literature for computing geometric moments, which utilise either lower-order approximating meshes or high-order spline surfaces representing $\mathcal{G}$. The most commonly employed method leverages Gauss's divergence theorem, which is also used in this work. This theorem facilitates the evaluation of moments by converting volume integrals to integrals over the surface bounding the volume; for further details, the interested reader may refer to~\cite{gmdsa_r5}.

\subsubsection{Relation of geometric moments to physics} As mentioned earlier, our motivation to include geometric moments in GOs stems from their extensive use in shape interrogation and their importance in a wide range of engineering analyses. In this subsection, we provide simple examples of such dependence for the case of hydrostatic and hydrodynamic characteristics via the use of the Sectional Area Curve (SAC) and its moments which are commonly used in computer-aided ship hull design. 
    
SAC is described by a function, $S(x)$, corresponding to the value of the submerged cross-sectional area (zeroth-order 2D geometric moment) of the ship at the longitudinal position $x$. In other words, it describes the longitudinal variation of the hull area below the waterline, thereby providing a quantified description of an important geometric property of the ship hull. This distribution is closely related to the ship hull's resistance and propulsion performance, as highlighted in~\cite{gmdsa_r62} where authors emphasise the decisive effect of SAC's shape on the global hydrodynamic properties of the hull. At the same time, one of the basic methods used in hull shape modification is based on SAC transformations, which have been globally adopted for ship-hull design and optimisation; see~\cite{gmdsa_r61}. 

Furthermore, linear wave-resistance studies, performed in \cite{gmdsa_r66,gmdsa_r68} and other works, have underlined the importance of SAC's longitudinal rate of change, i.e., $S'(x)$, which determines the strength of the Kelvin-source distribution used to model the disturbance caused by the body (ship) as it moves on the free-surface. Note that the flow around a slender hull cruising at a constant velocity on a free surface can be modelled by an appropriate source-sink distribution along its centerline. The strength of these sources is proportional to the longitudinal rate of change of the hull's cross-sectional area~\cite{gmdsa_r66, gmdsa_r68}. This aspect can be effectively captured by geometric moments, particularly those of higher order. In fact, Vosser's integral, an early derivation for evaluating the wave resistance coefficient ($C_w$) of slender hulls, explicitly depends on the longitudinal derivative of the cross-sectional area~\cite{gmdsa_r68}. 

If we now consider, the first derivative of $S(x)$, i.e., $S'(x) = \frac{dS(x)}{dx}$, where $S(x) = \int_{\Omega(x)} dydz$ and $\Omega(x)$ denote the submerged cross-sectional region of the hull at the longitudinal position $x$, we can easily compute the $p^{th}-$order moment of $S'(x)$, $m_p$ as follows:
\[
m_p=\int_0^Lx^pS'(x)dx = \left[x^pS(x)\right]_0^L-\int_0^Lpx^{p-1}S(x)dx,
\]
where $x=0$, $x=L$ correspond to the bow and stern tip positions, respectively. If we further assume that $S(0)=S(L)=0$, which is the common case for most ship hull, we get:
\begin{equation}\label{gmdsa_e29}
    m_p = -p\int_0^L x^{p-1}S(x)dx = -p \int_0^L\int_{\Omega(x)}x^{p-1}dxdydz,
\end{equation}
\noindent which finally leads to 
\begin{equation}\label{gmdsa_e30}
m_p = -pM^{p-1,0,0},
\end{equation}
where $M^{p-1,0,0}$ is the first component of $\bm{M}^s$ with $s=p+q+r=p-1$; see Eq. \eqref{eq:Mpqr}. Thus, the 1D $p^{th}-$order moments of $S'(x)$ are directly linked to the corresponding 3D $(p-1)^{th}-$order geometric moment of the hull. 
    
Obviously, not every physical quantity of interest is strongly connected with the included moments and the usage of $\mathcal{M}$ may only capture certain physics-related quantities. However, even when a strong connection to specific physics quantities is absent, geometric moments can still provide high-level intrinsic geometric information about the shape's geometry, as will be discussed and experimentally demonstrated later. This additional information is crucial for extracting efficient features, especially in DRM and GM, for producing design spaces with enhanced diversity and geometric validity.

\subsection{Curvature}
Curvature is a fundamental concept in shape analysis (differential geometry), which locally quantifies the rate a curve deviates from a straight or the rate a surface deviates from a plane. For curves, curvature is an intrinsic (embedding-independent) property, while Gaussian curvature, which is the product of the two principal curvatures, $\kappa_1\kappa_2$, is an intrinsic property of the surface. Total curvature, which in the case of surfaces corresponds to the integral of its Gaussian curvature, aggregates the local curvature information across parts or the entire surface, offering a powerful tool for understanding the geometric and topological properties of the shape. For example, the total curvature of a geodesic triangle on a surface equals the deviation of the sum of its angles from $\pi$, which offers a qualitative (positive/dome-like or negative/saddle-like) and quantitative measure of the shape's deviation from the plane. Even more fundamentally, the Gauss-Bonnet Theorem connects total curvature with the Euler characteristic, which is an intrinsic topological invariant~\cite{fu2008shape}.
    
Curvature is a valuable tool in various fields. In computer graphics (CG) and computer-aided geometric design (CAGD), it is employed as a general tool for assessing shape's quality and improving its fairness~\cite{hildebrandt2004anisotropic}. In Robotics, it can aid in planning smooth collision-free trajectories by avoiding sharp turns~\cite{chen2024fdspc}. In fluid dynamics, it is linked to fluid flow behaviour around objects~\cite{shen2017computational}. For example, the viscous-pressure resistance, expressed as an integral over the hull, depends on local surface properties like smoothness and curvature, with surface anomalies or roughness triggering flow separation and acting as turbulence generators. 

\subsubsection{Evaluation of Gaussian and total curvature}
We assume here that $\mathcal{G}$ is represented by a regular parametric surface $\mathbf{P}(u, v) = [x(u, v), y(u, v), z(u, v)]^T$, with its parameters, $u$ and $v$, lying in a planar domain $\mathcal{D}=[a,b]\times[c,d]\subset\mathbb{R}^2$. The Gaussian curvature, $K(u,v)=\kappa_1\kappa_2$, of $\mathbf{P}(u, v)$ characterises the behaviour of the surface in the neighbourhood the parametric point $(u,v)$, and can be calculated as the ratio of the determinants of the second (2FF) and first (1FF) fundamental forms:
    
    \begin{equation}
        K(u,v) = \kappa_1\kappa_2=\frac{\det(\text{2FF})}{\det(\text{1FF})}=\frac{LN - M^2}{EG - F^2},
    \end{equation}
    
    \noindent where, in matrix form, $\text{1FF} = \left[\begin{array}{cc}E & F\\F &G\end{array}\right],\,
        \text{2FF} = \left[\begin{array}{cc}L & M\\M &N\end{array}\right],$ and
    
    \begin{equation}
        E=\mathbf{P}_{u}\cdot\mathbf{P}_{u},\,F=\mathbf{P}_{u}\cdot\mathbf{P}_{v},\,G=\mathbf{P}_{v}\cdot\mathbf{P}_{v},\,L=\mathbf{n}\cdot\mathbf{P}_{uu},\,M=\mathbf{n}\cdot\mathbf{P}_{uv},\,\text\,N=\mathbf{n}\cdot\mathbf{P}_{vv},
    \end{equation}

\noindent with $\mathbf{n}(u,v)=\frac{\mathbf{P}_u \times \mathbf{P}_v}{|\mathbf{P}_u \times \mathbf{P}_v|}$ denoting the parametric normal vector at $(u,v)$ and the subscripts in $\mathbf{P}_{\bullet}$ denote the surface partial derivatives with respect to $\bullet$. Gaussian curvature $K(u,v)$ is an intrinsic property of the surface and provides valuable information about the local behaviour of the surface. Obviously, calculating the distribution of $K(u,v)$ on the surface provides information of the overall shape while its integral, i.e., total curvature, provides an overall shape-behaviour measure and a connection to its topology. Specifically, total curvature $\mathcal{K}$ corresponds to the integral:
    \begin{equation}\label{eq:totalcurvature}
        \mathcal{K}(\mathcal{G}) = \iint_\mathcal{G}KdA=\iint_{\mathcal{D}} K(u, v) | \mathbf{P}_u \times \mathbf{P}_v|\, dudv = \iint_\mathcal{D} K(u, v) \sqrt{EG-F^2}\, dudv,
    \end{equation}
    
\noindent where $dA$ is the area measure of $\mathcal{G}$. Alternatively, the Gauss–Bonnet Theorem can be used for calculating $\mathcal{K}(\mathcal{G})$ stating that for a compact, two-dimensional Riemannian manifold $\mathcal{G}$, the sum of $\mathcal{K}(\mathcal{G})$ and the integral of the geodesic curvature $\kappa_g$ along its boundary, $\partial\mathcal{G}$, parameterised by the arc length parameter $s$, is equal to $2\pi$ times the Euler characteristic $\chi(\mathcal{G})$, i.e., using Eq.~\eqref{eq:totalcurvature}
\begin{equation}
    \mathcal{K}(\mathcal{G}) = \iint_{\mathcal{D}} K(u, v) \sqrt{EG-F^2}\, dudv + \int_{\partial\mathcal{G}} \kappa_g(s) ds = 2\pi \chi(\mathcal{G}).
    \end{equation}
    
\noindent For a surface without a boundary, such as a closed surface, the boundary term $\int_{\partial\mathcal{G}} \kappa_g(s) ds$ is zero, and the theorem simplifies to:
\begin{equation}
    \mathcal{K}(\mathcal{G}) = \iint_{\mathcal{D}} K(u, v) \sqrt{EG-F^2}\, dudv = 2\pi \chi(\mathcal{G}).
\end{equation}

Gauss-Bonnet theorem connects the Gaussian curvature of the surface with the geodesic curvature of its boundary with the Euler characteristic, which is a topological invariant of the surface. This result is obviously surprising as regardless of how curvature at different regions changes its total curvature remains unchanged. In this paper, $\mathcal{K}$ is evaluated using the approach proposed in \cite{fu2008shape}, which estimates $\mathcal{K}$ on a triangulation approximating $\mathcal{G}$.

\subsubsection{Relation of curvature to physics}
In pertinent literature, various theoretical and numerical studies have demonstrated how variations in surface and curve curvature can impact the performance of resulting shapes. Particularly, these effects have been extensively explored in turbomachinery blades, wings, hydrofoil and aerofoil design. 
    
For instance, in the case of aerofoils, Shen et al.~\cite{ shen2017computational} utilised the Reynolds Averaged Navier–Stokes (RANS) method to investigate numerically the effects of curvature on boundary layer behaviour. They illustrated its significant influence on aerofoils’ aerodynamic performance. Additionally, they revealed that a discontinuous curvature distribution of an aerofoil can lead to a larger laminar separation bubble at lower angles of attack and lower Reynolds numbers, thereby affecting the aerofoil's performance at higher Reynolds numbers. Furthermore, Nemnem et al. \cite{nemnem2014smooth} developed a fifth-order mean-line by twice integrating a cubic B-spline to shape the aerofoil. This ensured continuous curvature distributions, thereby enhancing the performance of corresponding turbine blades. A design approach, known as the prescribed surface curvature distribution blade design (CIRCLE) method, was proposed by Korakianitis and Papagiannidis~\cite{korakianitis1992surface} to optimise aerofoils and blades by ensuring continuous curvature and slope-of-curvature distribution along their surfaces. This method highlighted the significant impact of surface curvature distribution on aerodynamic and heat transfer performance, highlighting the importance of maintaining curvature continuity for enhanced blade performance. Korakianitis et al. \cite{10.1115/1.4005969} applied the CIRCLE method to two wind turbine aerofoils, concluding that higher continuity curvature distributions improved the aerodynamic performance compared to the original wind turbine aerofoils. Massardo et al. \cite{massardo1989axial_I,massardo1989axial_II} employed streamline curvature distribution calculations to determine the 3D variation of inlet and outlet flow angles in axial-flow compressor design, thereby improving compressor efficiency. Similarly, Song and Gu \cite{song2014effects} demonstrated the importance of improving curvature continuity at the leading-edge blend point of a compressor blade, which aided in eliminating the separation bubble and enhancing blade performance, underscoring the significance of maintaining curvature continuity.

\subsection{Three-dimensional Fourier descriptor}
Similar to geometric moments, Fourier Descriptors (FDs) are widely used in 2D and 3D shape analysis tasks, such as  object recognition, reconstruction and classification both in traditional approaches and modern ML methods for similar tasks. However, the two approaches differ in the shape properties they describe. Specifically, while geometric moments capture volume distributions of a 3D shape, FDs are particularly sensitive to the geometry of the shape's boundary~\cite{reeves1988three}. For example, the FDs for a \textit{slotted aerofoil} with a blunt trailing edge and an \textit{integral aerofoil} (i.e., an aerofoil with a continuous surface without any slots or gaps) will vary significantly due to their different shapes, although their moment-based representations might be similar. Therefore, the inclusion of FDs in the proposed GOs enables an enhanced description of the object's boundary surface, which is a critical factor in functional surfaces (wings, blades, ship hulls, etc.) considered in this work. A brief description of the evaluation of 3D FDs, based on the work of Park and Lee~\cite{park1987three} is presented subsequently. Although several approaches are available in the pertinent literature, the method proposed in~\cite{park1987three} is a simple, but sufficiently accurate approach in the context of the proposed GOs, which is briefly described in sequel.  

Consider $\mathbf{C}_z(s) = X_z(s)+iY_z(s) \in \mathcal{G}$ being a complex function that represents a continuously varying boundary of parallel cross sections of the closed surface $\mathcal{G}$, lying on planes parallel to the $xy$-plane. Each such section is at a height $z$ and has an arc-length parametrization $s$. As we have assumed that $G$ is a closed surface, $\mathbf{C}_z(s)$  ($z=$constant) will be a closed periodic curve on s, i.e., $\mathbf{C}_z(s+nL_z) = \mathbf{C}_z(s)$, where $n\in\mathbb{N}_0$ and $L_z$ is the section's length at height $z$. We may now compute its Fourier coefficients $\mathbf{F}(z, n)$ as

\begin{equation} \label{fd_1}
\mathbf{F}(z, n) = \frac{1}{L_z} \int_0^{L_z} \mathbf{C}_z(s)e^{\frac{-i2\pi n s}{L_z}}\,ds,\quad  n\in\mathbb{Z}.
\end{equation}

\noindent For a fixed height $z$, the $\mathbf{F}(z, n)$ coefficients are equivalent to the planar FD and represent amplitudes of harmonic frequencies (power spectrum) of the section $\mathbf{C}_z(s)$ ($z=$constant).  Given $\mathbf{F}(z, n)$, the cross sections can be reconstructed via the inverse Fourier transformation:
\begin{equation}\label{fd_2}
    \mathbf{C}_z(s) = \sum_{n=-\infty}^{\infty} \mathbf{F}(z, n)e^{\frac{i2\pi n s}{L_z}}.
\end{equation}
    
\noindent If we now want to capture shape variation along the z-axis, another Fourier transform (of $\mathbf{F}(z, n)$ this time) with respect to $z$ is needed. If we assume that $z\in[0,H]$, the resulting Fourier coefficients become:

\begin{equation}\label{fd_3}
\mathcal{F}(m, n) = \frac{1}{H} \int_0^H \mathbf{F}(z, n)e^{\frac{-i2\pi m z}{H}}\,dz, \quad m\in\mathbb{Z}.
\end{equation}

\noindent with the former Fourier coefficients being reconstructed, as before, via the inverse Fourier transformation:

\begin{equation}\label{fd_4}
\mathbf{F}(z, n) = \sum_{m=-\infty}^{\infty} \mathcal{F}(m, n)e^{\frac{i2\pi m z}{H}}.
\end{equation}
    
\noindent Collectively, if we substitute Eq.~\eqref{fd_1} into Eq.~\eqref{fd_3}, we obtain the Fourier transform of $\mathbf{C}_z(s)$ as:

\begin{equation}\label{fd_5}
\mathcal{F}(m, n) = \frac{1}{H} \int_0^H \left( \frac{1}{L_z} \int_0^{L_z} \mathbf{C}_z(s)e^{\frac{-i2\pi n s}{L_z}}\,ds \right) e^{\frac{-i2\pi m z}{H}}\,dz, \quad m,n\in\mathbb{Z}.
\end{equation}

\noindent If we consider that $\mathcal{G}$ is discretised into a finite number of parallel sections, each one of them further discretised into a number of points, the corresponding integrals in Eqs.~\eqref{fd_1},\eqref{fd_3} and \eqref{fd_5} are approximated by summations leading to a double discrete Fourier transformation. From a shape analysis perspective, we can say that the information about the original shape is encoded in the magnitudes of its Fourier coefficients, i.e., $\mathcal{F}(m, n)$. Larger magnitudes in specific coefficients indicate that those particular frequencies play a more significant role in defining $\mathcal{G}$'s characteristics.

Adding $\mathcal{F}(m, n)$ to GOs would drastically increase the size of the feature vector, as a large number would be required for complicated shapes. However, $\mathrm{GO}(\mathcal{G})$ in Eq.~\eqref{gmdsa_e29} comprises multiple operators and each component may only capture parts of the shape's geometry and characteristics. Therefore, in this work, we make use of Parseval's theorem, which, in the context of signal theory, states that the sum of the squares of the magnitudes of the Fourier coefficients is equal to the total energy of the original signal. In shape analysis, we can consider the original shape as a signal and its Fourier coefficients as capturing its geometric variations.

Hence, we can rephrase Parseval's theorem by stating that the sum of the squared magnitudes of the Fourier coefficients represents the total variance of the original shape. Here, variance refers to how much the shape' boundary deviates from its mean form.

Therefore, we may state that the information in $\mathcal{G}$ is preserved through its total energy (or total variance associated with the geometry of the shape's boundary), which is the sum of the squared magnitudes of its Fourier coefficients, and can be represented as follows: 
\begin{equation}
\mathcal{F}_T(\mathcal{G}) = \int_{-\infty}^{\infty} \int_{-\infty}^{\infty} |\mathcal{F}(m, n)|^2 \, dmdn,
\end{equation}
with, finally, $\mathcal{F}_T(\mathcal{G})$ being the component which substitutes $\mathcal{F}(\mathcal{G})$ in Eq.~\eqref{GO_1}.

\section{Experiments: Test cases, models' training and analyses}
To validate the utility of GOs in model training, we selected two distinct design cases: 2D aerofoil profiles and 3D ship hull surfaces, which are of varying dimensions and complexity. At the same time, both cases involve free-form geometries and significant computational costs for the assessment of their performance. This obviously brings significant challenges when design optimisation is considered. Furthermore, each case requires its own set of approaches when considering appropriate geometric representations and the physical quantities that are of interest in each of them. In each case, a set of design problem variations is considered (a common occurrence in the industry) which highlights the flexibility of the proposed methodology in adapting to varying requirements, while the twofold comparison enables us to demonstrate its versatility in addressing distinctly different engineering problems. We leveraged both test cases to showcase the effectiveness of GOs in supporting the training of SMs, GMs and DRMs. This section is devoted to presenting a systematic experimentation with detailed analysis that helps us in assessing the utility of integrating GOs in the respective models when aiming to enhance their performance.

\subsection{Training datasets}
The development of SMs, GMs or DRMs is preceded by the compilation of appropriate datasets comprising sufficiently diverse design instances. These instances are retrieved from pertinent design resources and/or are, quite commonly, automatically generated by parametric models for both aerofoils and ship hulls. These parametric models and the corresponding datasets will be described in sequel as they are integral to the training of SMs, GMs and DRMs,

\subsubsection{Ship hull test case}
\begin{figure*}[hbt!]
\centering
\includegraphics[width=0.9\textwidth]{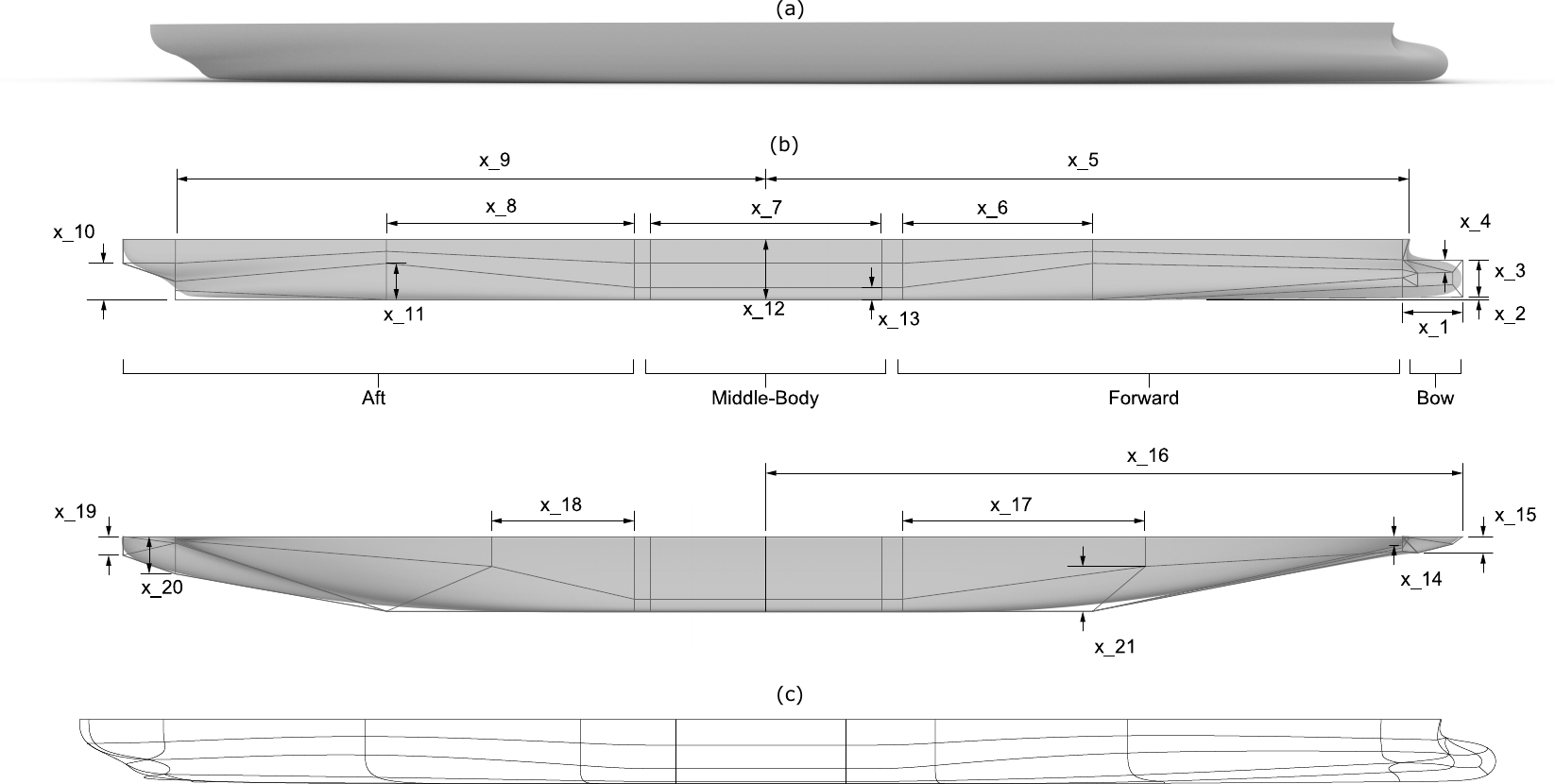}
\caption{(a) The 3D surface model, (b) its parameterisation, and (c) the resulting surface patches used in numerical evaluations.}
\label{op_7}
\end{figure*}

The ship hull dataset generation was performed via the parametric model presented in~\cite{gmdsa_r21}, which was originally designed to parametrically generate container ship hulls, but was later extended to cover a wide range of ship types. This model uses a relatively small set of parameters that determine component properties of a high-level construction scheme that procedurally generates the corresponding ship hull surface instance. One such ship hull instance, closely resembling the well-known KCS\footnote{\url{https://www.nmri.go.jp/study/research_organization/fluid_performance/cfd/cfdws05/gothenburg2000/KCS/kcs_l&r.htm}} hull, is shown in Figure~\ref{op_7}(a). The employed high-level construction scheme controls the overall hull shape, transition areas, as well as local free-form features by determining the position of control points of a T-spline surface via linear and nonlinear relations. This approach enables a comprehensive feature-driven parameterisation, where each parameter influences a distinct aspect of the hull's form, facilitating both local and global modifications while guaranteeing the generation of valid ship hull designs with no self-intersection or other unwanted behaviours. The parameters in the design vector, denoted by $\mathbf{x}$, are categorised into global, local, and transitional types, each affecting the hull shape differently. An indicative portrayal of geometric features and the corresponding parameters that control their shape is included in Figure~\ref{op_7}(b).
    
Global parameters, such as the ship's length (at the waterline), beam, and depth, are the most influential in terms of the overall dimensions and the global ship form. Typically, these parameters are predefined during the conceptual design phase, based on customer requirements, and therefore, they are either kept fixed or only small variations are allowed. Local parameters, such as those defining dimensions of different parts of the flat of side or flat of bottom, as well as the bilge radius, shaft dimensions, and the bulbous bow and stern affect smaller areas of the ship hull, but nevertheless significantly modify its performance characteristics~\cite{gmdsa_r21}. Ship hull regions, such as the midship part, or transitional areas, e.g., from mid-hull to bow, are local in nature, but still affect a significant part of the overall shape and are therefore considered as a separate category in this context. By ignoring some global, commonly fixed, parameters, as well as keeping constant values for local parameters that do not affect the ship hull's performance, we can achieve a hull instance generation process with only 21 non-dimensionalised parameters, $\mathbf{x}=\lbrace x_1,x_2,x_3\dots x_{21}\rbrace$,  $x_i\in[0,1]\, \forall i$, that produce plausible and geometrically valid shape modifications while allowing the full range of possible variations to performance-sensitive areas of the ship hull. 
    
With respect to the performance criterion, the wave-making resistance, $R_w$, and specifically its corresponding coefficient, $C_w$, was employed in this study. The wave-making resistance is a part of the total hull resistance, $R_T$, which is the overall force exerted on the moving body by the fluid. Wave-making resistance corresponds to the energy spent on creating free-surface waves by the moving surface watercraft and it is not to be confused with the wave-breaking resistance which corresponds to the resistance caused by preexisting surface waves. $R_T$ comprises various components, with viscous resistance, $R_V$, typically constituting a major part of it, however, $R_w$ becomes a major component for displacement hulls travelling at relatively high speeds. It is also noteworthy that $C_w$ is highly sensitive to local hull features and therefore significant $R_w$ can be achieved without affecting the overall shape and cargo capacity~\cite{khan2022geometric}. $C_w$ is influenced by the distribution of the hull's shape, similar to some GOs, and can be used as a physics-informed shape signature when considering shape representations. Minimising $R_w$ at the preliminary design stage is crucial for certain types of ship hulls, but its evaluation can be computationally demanding; therefore, creating a surrogate for $C_w$ is essential, especially during optimisation.

In this work, the estimation of $C_w$ was performed using an Isogeometric Analysis-based Boundary Element Method (IGA-BEM), initially presented by Belibassakis et al.~\cite{gmdsa_r22} for NURBS surface representations, and later extended to also handle T-splines in~\cite{GINNIS2014425,gmdsa_r21}. This solver employs Isogeometric Analysis (IGA) to solve the boundary integral equation (BIE) associated with the linearised Neumann–Kelvin formulation, which is essential for calculating $C_w$. The T-spline elements used for the evaluation of the velocity potential and subsequently $C_w$ are illustrated in Figure \ref{op_7} (c).
    
\subsubsection{Aerofoil test case}
For the case of aerofoils, two distinct datasets are considered. The first dataset is generated parametrically, akin to the previously described ship hull scenario, while the second is derived from the publicly available database of foil designs at UIUC\footnote{\url{https://m-selig.ae.illinois.edu/ads/coord_database.html}}. These datasets correspond to different modalities since the first one contains parameter vectors for aerofoil representations, whereas the second one comprises discretised aerofoils represented as lists of coordinates of the corresponding point sets. This twofold approach is required to accommodate the training needs of GMs, but it also provides us with opportunity to demonstrate the efficacy of GOs across varying data modalities.

\begin{figure*}[hbt!]
\centering
\includegraphics[width=0.9\textwidth]{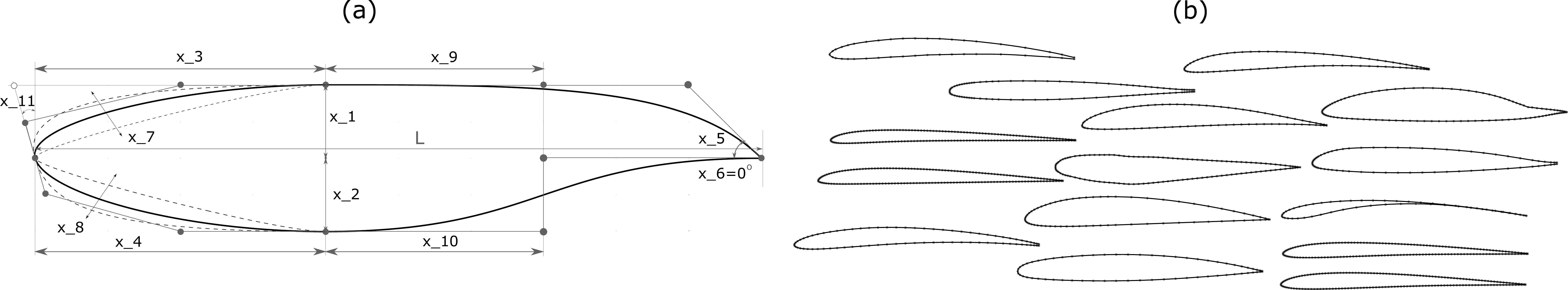}
\caption{(a) Parameterisation of aerofoil profiles. (b) Randomly selected samples from the UIUC aerofoil database.}
\label{aero_param}
\end{figure*}

For the parametric generation of aerofoils, the standard reference model from the family of parametric models described in~\cite{KOSTAS201779} has been used. This standard model comprises 11 non-dimensionalised parameters which are depicted in Figure~\ref{aero_param}(a). The correspond design vectors, comprising obviously 11 components, were used to delineate each aerofoil's profile, and the resulting dataset was used in training of both SMs and DRMs. The parametric aerofoil construction begins with the definition of four simple cubic B\'ezier curves, which are then combined to form the aerofoil profile described by a single cubic B-spline curve. The aerofoil's chord length is used for normalisation, with the remaining length parameters assuming a value in $[0,1]$. Therefore, in this case, $\mathcal{P}(\mathcal{G})=\lbrace x_i,x_2,x_3\dots x_{11}\rbrace$ constitutes the design vector that generates a unique profile for each combination of $\lbrace x_i\rbrace_{i=1}^{11},\,x_i\in[0,1]\,\forall i$, values. Note that this standard parametric model can accurately reconstruct, i.e, within wind-tunnel tolerance~\cite{doi:10.2514/1.29958}, approximately 75\% of the aerofoils in the UIUC database.
    
As mentioned before, GMs are trained with the dataset used in~\cite{chen2020airfoil}, which contains 1,600 real-world airfoil designs from the UIUC database (see Figure~\ref{aero_param}(b)), along with their synthetically created variations. Each design in this training set was pre-processed to achieve a uniform cardinality of 192 2D discrete points along its suction (upper) and pressure (lower) sides. In this case, $\mathcal{P}(\mathcal{G})$ contains the coordinates of these points, i.e., $\mathcal{P}(\mathcal{G}) = \lbrace x_1, x_2, x_3, \dots, x_{384}\rbrace$, resulting in a design space with 384 dimensions. 

The lift-to-drag ratio, as expressed by the corresponding coefficients, i.e., $C_L/C_D$, serves as the primary performance metric in this design case. Although an IGA-BEM approach~\cite{KOSTAS201779}, similar to the one used for the ship hull case, exists for the calculation of the lift coefficient, this approach is based on potential theory and therefore we cannot obtain drag coefficient estimates directly. Hence, the well-known XFOIL computational package~\cite{Drela87,Drela89} was used instead.  Performance evaluations were carried out at a fixed angle of attack, set to $3^\circ$, and a Reynolds number, Re, of $5\times10^5$ corresponding to a Mach number, Ma, of approximately $2\times10^{-2}$, when aerofoils are considered, or approximately $4\times10^{-4}$ if hydrofoils were to be considered instead. 

\subsection{Discriminative model training}\label{DTR_DM}
Neural Networks (NNs)~\cite{iliadis2023engineering} and Gaussian Process Regression models (GPRs)~\cite{goodfellow2016deep} were used as surrogate models for the case of SMs, which were also employed for testing and validating the performance of the proposed GOs. Specifically, in both cases, i.e., NNs and GPRs, the dependent variables in the training dataset corresponded to $\mathcal{P}$, which in this case composed of design parameters, along with varying combinations of the remaining GOs, i.e., geometric moments ($\mathcal{M}$), total curvature ($\mathcal{K}$) and total energy/variance ($\mathcal{F}_T$). As for the independent variable, $C_w$ and $C_L/C_D$ were used for the ship hull and aerofoil design case, respectively. Both the architecture and hyper-parameters of the employed NNs and GRPs were optimised before use in this study. For NNs, the number of hidden layers ranged from 2 to 6 with 2 to 30 neurons in each layer. For GPR models, we examined different types of kernels, including linear, polynomial, periodic, Mat\'ern, radial basis function, and rational quadratic kernels, with both constant and linear mean functions~\cite{piffl_r23}. The Mean Square Error (MSE) was minimised when training NNs while the negative log-likelihood loss function was used for GPRs. Finally, model training was performed using the Adam optimiser, with varying values of its hyper-parameters, including weight decay $=[0, 10^{-4}, 10^{-5}, 10^{-6}]$, $\beta_1=[0.9, 0.95]$, $\beta_2=0.999$, and learning rates in $[10^{-1}, 10^{-2}, \dots, 10^{-6}]$. For both the parametric hull and aerofoil cases shown in Figure \ref{op_7}(b) and \ref{aero_param}(a), we sampled 11- and 21-dimensional design spaces, respectively, to create $n=500$ distinct designs, i.e., $\mathcal{X} = \lbrace \mathbf{x}_1, \mathbf{x}_2, \dots, \mathbf{x}_{500} \rbrace$. For sampling, we used the STLBO approach~\cite{khan2018sampling}, which ensures that all designs are evenly distributed over the entire design space defined by the corresponding datasets mentioned earlier. From dataset $\mathcal{X}$ , 80\% and 20\% of the designs were used for model training and validation, respectively.

To assess the effect of GOs on the models' performance, we trained them using the design parameters $\mathcal{P}$, and varying combinations of the integral operators, i.e., $\left(\mathcal{P},\mathcal{M}\right)$, $\left(\mathcal{P},\mathcal{K}\right)$ $\left(\mathcal{P},\mathcal{F}_T\right)$, $\left(\mathcal{P},\mathcal{M},\mathcal{K}\right)$, $\left(\mathcal{P},\mathcal{M},\mathcal{F}_T\right)$, $\left(\mathcal{P},\mathcal{K},\mathcal{F}_T\right)$, and   $\left(\mathcal{P},\mathcal{M},\mathcal{K},\mathcal{F}_T\right)$. For example, when calculating $\mathcal{M}$ for ship hulls (see Eq.~\eqref{eq:SSV}), we included moments up to  $s = 5$ order, resulting in $n_\mathcal{M} = 56$ (see Eq. \eqref{eq:SSVcardinality}) moment components, which were sufficient for capturing both global and local geometric features. In Table~\ref{table_2}, we report the values of these components for the hull instance shown in Figure~\ref{op_7}. For the same hull instance, the remaining integral operators, i.e., $\mathcal{K}$ and $\mathcal{F}_T$, assume the values of $6.45\times10^9$ and $93.15$, respectively, Note that in the second design case, i.e., aerofoils, we have a two-dimensional design and consequently, $\mathcal{M}$ will only include geometric moments with respect to two axes. Hence Eq.~\eqref{mf} can be rewritten as $\bm{M}^s = \{M^{p,q} \mid p+q = s\}$. For this 2D case, we included moments up to $s = 10$ which results in $n_\mathcal{M} = 62$ components.

\begin{table}[htb!]
    \small
    \centering
    \caption{Values of $\mathcal{M}$ for the ship hull instance shown in Figure~\ref{op_7}.}
    \begin{tabular}{*{7}{c}}
     \toprule
     \multicolumn{7}{c}{$\mathcal{M}$}  \\
     \toprule
     $\bm{M^{0,0,0}}$ & $\bm{M^{0,0,1}}$ & $\bm{M^{0,1,0}}$ & $\bm{M^{1,0,0}}$ & $\bm{M^{0,0,2}}$ & $\bm{M^{0,1,1}}$ & $\bm{M^{0,2,0}}$  \\
     $8.77\times 10^{4}$ & $-5.48\times 10^{5}$ & $1.46 \times 10^{-11}$ & $-1.23\times 10^{7}$ & $4.61\times 10^{6}$ & $1.16 \times 10^{-10}$ & $6.04\times 10^{6}$  \\
     \hline
     $\bm{M^{1,0,1}}$ & $\bm{M^{1,1,0}}$ & $\bm{M^{2,0,0}}$ & $\bm{M^{0,0,3}}$ & $\bm{M^{0,1,2}}$ & $\bm{M^{0,2,1}}$ & $\bm{M^{0,3,0}}$ \\
     $7.57\times 10^{7}$ & $5.59 \times 10^{-9}$ & $2.09\times 10^{9}$ & $-4.37\times 10^{7}$ & $0.0$ & $-3.74\times 10^{7}$ & $0.0$  \\
     \hline
     $\bm{M^{1,0,2}}$ & $\bm{M^{1,1,1}}$ & $\bm{M^{1,2,0}}$ & $\bm{M^{2,0,1}}$ & $\bm{M^{2,1,0}}$ & $\bm{M^{3,0,0}}$ & $\bm{M^{0,0,4}}$  \\
     $-6.33\times 10^{8}$ & $1.49 \times 10^{-8}$ & $-8.57\times 10^{8}$ & $-1.27\times 10^{10}$ & $4.77 \times 10^{-7}$ & $-3.97\times 10^{11}$ & $4.45\times 10^{8}$ \\
     \hline
     $\bm{M^{0,1,3}}$ & $\bm{M^{0,2,2}}$ & $\bm{M^{0,3,1}}$ & $\bm{M^{0,4,0}}$ & $\bm{M^{1,0,3}}$ & $\bm{M^{1,1,2}}$ & $\bm{M^{1,2,1}}$ \\
     $0.0$ & $3.12\times 10^{8}$ & $0.0$ & $8.39\times 10^{8}$ & $6.02\times 10^{9}$ & $0.0$ & $5.25\times 10^{9}$  \\
     \hline
     $\bm{M^{1,3,0}}$ & $\bm{M^{2,0,2}}$ & $\bm{M^{2,1,1}}$ & $\bm{M^{2,2,0}}$ & $\bm{M^{3,0,1}}$ & $\bm{M^{3,1,0}}$ & $\bm{M^{4,0,0}}$ \\
     $0.0$ & $1.05\times 10^{11}$ & $0.0$ & $1.39\times 10^{11}$ & $2.36\times 10^{12}$ & $0.0$ & $8.07\times 10^{13}$ \\
     \hline
     $\bm{M^{0,0,5}}$ & $\bm{M^{0,1,4}}$ & $\bm{M^{0,2,3}}$ & $\bm{M^{0,3,2}}$ & $\bm{M^{0,4,1}}$ & $\bm{M^{0,5,0}}$ & $\bm{M^{1,0,4}}$ \\
     $-4.72\times 10^{9}$ & $0.0$ & $-2.94\times 10^{9}$ & $0.0$ & $-5.17\times 10^{9}$ & $0.0$ & $-6.12\times 10^{10}$ \\
     \hline
     $\bm{M^{1,1,3}}$ & $\bm{M^{1,2,2}}$ & $\bm{M^{1,3,1}}$ & $\bm{M^{1,4,0}}$ & $\bm{M^{2,0,3}}$ & $\bm{M^{2,1,2}}$ & $\bm{M^{2,2,1}}$ \\
     $0.0$ & $-4.36\times 10^{10}$ & $0.0$ & $-1.19\times 10^{11}$ & $-9.93\times 10^{11}$ & $0.0$ & $-8.36\times 10^{11}$  \\
     \hline
     $\bm{M^{2,3,0}}$ & $\bm{M^{3,0,2}}$ & $\bm{M^{3,1,1}}$ & $\bm{M^{3,2,0}}$ & $\bm{M^{4,0,1}}$ & $\bm{M^{4,1,0}}$ & $\bm{M^{5,0,0}}$ \\
     $0.0$ & $-1.94\times 10^{13}$ & $0.0$ & $-2.46\times 10^{13}$ & $-4.71\times 10^{14}$ & $0.0$ & $-1.72\times 10^{16}$ \\
     \bottomrule
    \end{tabular}
    \label{table_2}
\end{table}

\begin{figure*}[hbt!]
    \centering
    \includegraphics[width=01\textwidth]{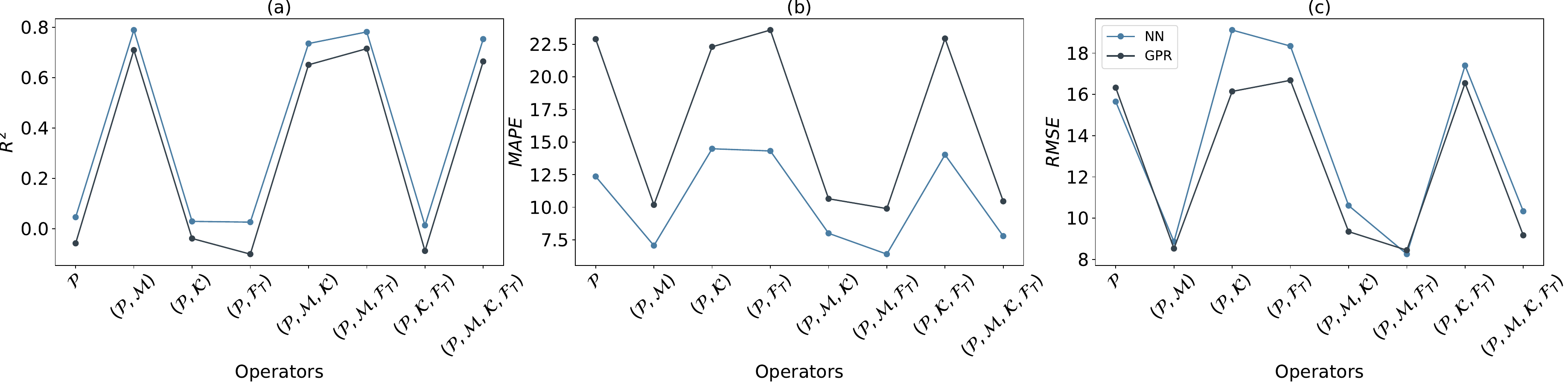}
    \caption{Plots of $R^2$, MAPE, and RMSE for various combinations of GOs for the optimised NNs and GPRs trained for aerofoils.}
        \label{aerofoil_train}
    \end{figure*}

\begin{figure*}[hbt!]
    \centering
\includegraphics[width=01\textwidth]{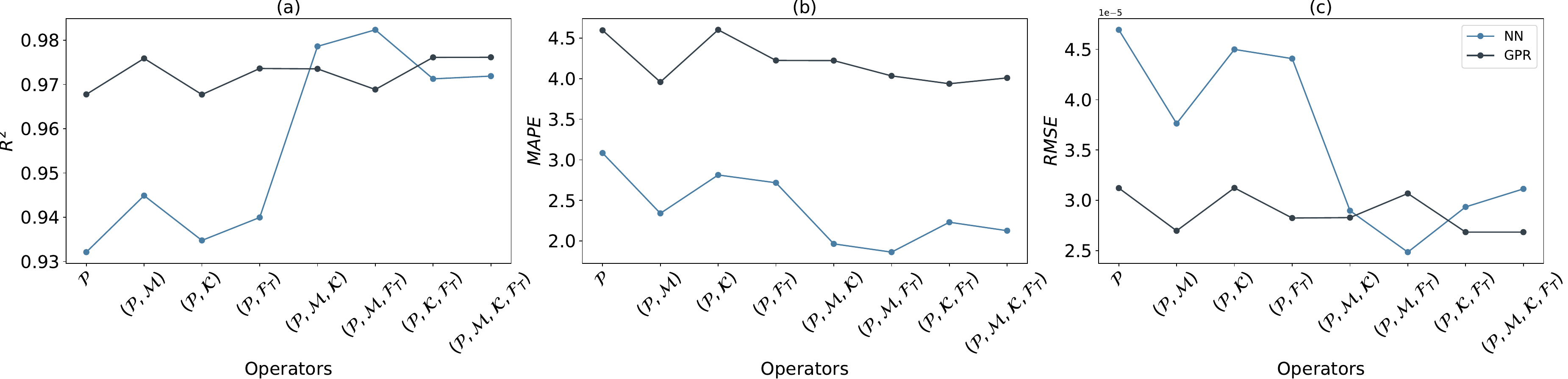}
    \caption{Plots of $R^2$, MAPE, and RMSE for various combinations of GOs for the optimised NNs and GPRs trained for ship hulls.}
    \label{hull_train}
\end{figure*}

To study the effect of GOs in surrogate modelling, the baseline scenario involves a model training using only $\mathcal{P}$, i.e., the design parameters of the hull and aerofoil, as the dependent variables. The training results of NNs and GPRs are shown in Figure~\ref{aerofoil_train} for the aerofoil case and in Figure~\ref{hull_train} for the ship hull design case. In both figures, the coefficient of determination, $R^2$, the mean absolute percentage error, MAPE, and the root mean squared error, RMSE, are plotted for all combinations of GOs. 

\textbf{Aerofoil surrogate model results:} As can be easily observed in Figure~\ref{aerofoil_train}, using only $\mathcal{P}$ yields poor results across all metrics and both training models, i.e., low $R^2$ values and relative large values for both MAPE and RMSE. The same picture is drawn when total curvature ($\mathcal{K}$) and/or total variance ($\mathcal{F}_T$) are included in the features vector. This observation indicates that these shape descriptors (and their combination) do not offer any advantage to either regression models and their predictive power with respect to the employed performance metric, $C_L/C_D$, is rather poor. However, this picture is completely overturned when geometric moments are introduced. The introduction of $\mathcal{M}$ to any descriptor combination significantly improves the performance metrics for both models, with a noticeable jump in $R^2$ and a reduction in both MAPE and RMSE values. This suggests that $\mathcal{M}$ is a highly predictive feature vector for the lift-over-drag ratio, $C_L/C_D$, and presumably highly correlated with its value. Overall, the highest level of improvements, for both regression models, was recorded for the shape descriptor comprising $\left(\mathcal{P},\mathcal{M},\mathcal{F}_T\right)$, without however being significantly better to $\left(\mathcal{P},\mathcal{M}\right)$. Both regression models, NN and GPR, reveal a similar trend across all shape descriptor combinations as seen in Figure~\ref{aerofoil_train}. However, neural networks have better performance with respect to $R^2$ and MAPE and only slightly worse for RMSE values. This suggests that the trained NN models are better in capturing the dataset variance compared to GPRs.

\textbf{Ship hull surrogate model results:} 
Figure~\ref{hull_train} depicts the same training results for the ship hull case. This time a different behaviour is exhibited across all metrics while a more pronounced difference is recorded between the two regression models. Overall, the wave resistance coefficient value is predicted with sufficient accuracy, even when only $\mathcal{P}$ is used. Specifically, if we compare Figures~\ref{hull_train} and \ref{aerofoil_train}, we observe an average MAPE value between 2\% and 4.5\% for $C_w$ which is significantly lower than the $C_L/C_D$ (between 7\% and 22.5\%) case observed before. Furthermore, the inclusion of each operator, on top of $\mathcal{P}$, enhances the model's ability to capture the complexity of the hull's wave resistance, and predicting the value of $C_w$. The effect of positive contributions with each added operator is easily observed for NNs while GPRs exhibit only minor improvements with each additional shape descriptor. The best descriptors, as revealed by Figure~\ref{hull_train}, are $\lbrace\mathcal{P},\mathcal{M},\mathcal{K}\rbrace$ and $\lbrace\mathcal{P},\mathcal{M},\mathcal{F}_T\rbrace$ for the case of NNs, while $\lbrace\mathcal{P},\mathcal{K},\mathcal{F}_T\rbrace$ and $\lbrace\mathcal{P},\mathcal{M},\mathcal{K},\mathcal{F}_T\rbrace$ are the best performers when GPR is used. Despite the general improvement trend, recorded in both regression models, and contrary to NN, the GPR model does not seem to get significant benefits from the proposed descriptors. Nevertheless, even such small improvements can still be beneficial at a later stage when the model is used during optimisation.

\subsection{Design space dimensionality reduction}
For design space dimensionality reduction (DR), we once again use GOs to augment the training data with high-level and compact design descriptions. As will be demonstrated later in this section, this augmentation provides a physics-informed design description, enabling the models to learn an efficient subspace characterised by a higher degree of diversity and validity compared to the subspace learned without the use of GOs. We demonstrate the performance of GOs in dimension reduction through subspace learning (feature extraction) and feature selection. For subspace learning, i.e., feature extraction, we employed both ML and non-ML-based approaches, with autoencoders (AEs)~\cite{goodfellow2016deep} being the selection from the former category, and Karhunen--Lo{\`e}ve expansion (KLE)~\cite{diez2015design} from the latter one. In both cases, the introduction of GOs enables these relatively simple models to capture detailed and compact geometric representations of the original designs in the latent space. 

For the employed AEs, both encoder and decoder architectures are based on NNs, with their training setup mirroring the one used in NN-based surrogate models in \S\ref{DTR_DM}. In a similar fashion, the architecture and hyper-parameters that yield the lowest MSE on the test dataset are selected for the final optimised AE model. Note that although more advanced AE models could have been used, the primary aim in this work is to enhance standard models with GOs so that they can achieve higher learning accuracy without having to resort to complicated and computationally costly procedures. The metrics used for evaluating learning accuracy include representational \textit{capacity} and \textit{compactness} which will be discussed in detail later in this section.

The KLE-based approach proposed by Diez et al.~\cite{diez2015design} solves a variational problem that produces an optimal orthogonal basis, which is used for the reduced-dimensionality linear representation of the original design space. The KLE features are characterised by the eigenfunctions of a symmetric and positive definite covariance function, which is constructed using GOs of the training dataset. The KLE values, associated with each feature, facilitate the differentiation between active and inactive features. The reparameterisation of the shape with active features serves as a new basis for defining subspaces that capture the highest variance in geometry. By setting a desired variance level, the corresponding number of features is utilised to construct a subspace with lower dimensionality when compared to the original design space. 

Finally, for feature selection, we apply sensitivity analysis (SA) using the Sobol approach~\cite{gmdsa_r32}. This method is widely used within the engineering community, as it provides a comprehensive overview of the sensitivity or significance of all parameters in relation to the physics employed in each case. As mentioned in \S\ref{INTRO}, parameters that are more sensitive to physics are deemed significant for improving designs' performance in the context of design optimisation. Consequently, those with minimal impact on performance are typically fixed or excluded from the optimisation study in order to reduce the dimensionality of optimisation problem.

\subsubsection{DR with subspace learning}
The primary aim of this approach is to reduce the dimensions of the design space by learning/extracting a relatively small number of latent features, which can be subsequently used in design operations that are hindered by high-dimensional spaces. The same design cases, i.e., ship hulls and aerofoils, are considered again. The quality of the resulting latent subspace is typically assessed on the basis of its representational capacity and compactness. These qualities, as defined in~\cite{chen2021padgan}, refer to the subspace's ability to produce diverse and valid shapes, respectively. A latent subspace producing invalid or inadequately diverse designs is obviously unsuitable for optimisation as the computational budget will be mainly spent on exploring infeasible or similar designs. At the same, when using feature extraction techniques, it is equally important to secure an appropriate shape representation, since otherwise the latent subspace may be lacking features needed for performance optimisation uses. This limitation mainly arises from low-level shape discretisations which are commonly used in classical subspace learning approaches.  Therefore, richer representations containing high-level particulars related to the shape's structure and physics, such as the ones embedded in the proposed GOs, are essential in the design optimisation setting.

\begin{figure*}[hbt!]
\centering
\includegraphics[width=01\textwidth]{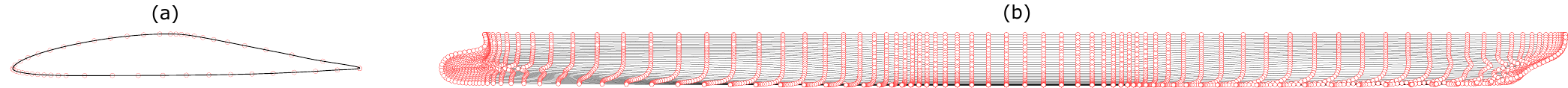}
\caption{Distribution of points on the (a) aerofoil and (b) hull test cases used for the training of KLE and AE.}
\label{design_dist}
\end{figure*}

Since geometry discretisations are essential for describing the low-level shape information in such methods, all designs, i.e., hull surfaces and aerofoil profiles, need a corresponding $\mathcal{P}$ descriptor of this type. In this work, we employ a 3D discretisation approach for ship hulls using the shape-encoded method proposed in~\cite{khan2023shiphullgan}, which translates all designs into a common template representation. This representation is composed of quadrilateral mesh elements, with the vector of coordinates of these mesh vertices constituting $\mathcal{P}$ in this case. For the aerofoil case, $\mathcal{P}$ comprises $xy$--coordinates of 192 profile points sampled according to the methods proposed in~\cite{masood2024generative}. Specifically, a curvature-based approach is selected for sampling which  ensures point concentration near regions of high curvature, such as the leading edge region. Figures~\ref{design_dist}(a) and \ref{design_dist}(b) illustrate discretisation examples for aerofoil and ship hull designs, respectively.

\subsubsection{Subspace Learning Results} 
\begin{figure*}[hbt!]
\centering        \includegraphics[width=01\textwidth]{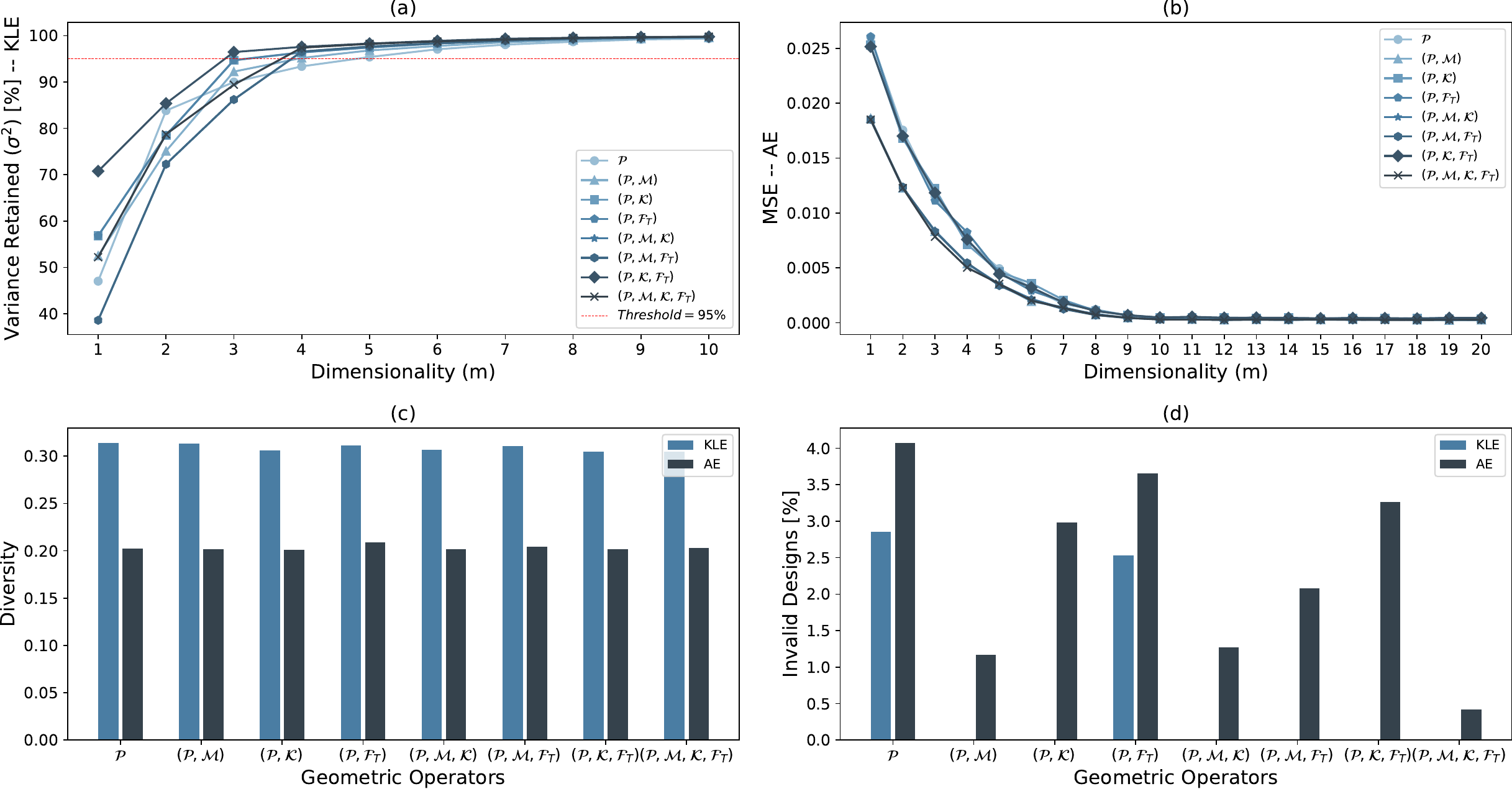}
\caption{Aerofoil Design Spaces: (a) retained percentage of variance with respect to KLE-based subspace's dimension; the horizontal red line indicates the 95\% threshold. (b) Reconstruction accuracy with respect to AE-based subspace's dimension measured via MSE. (c) Comparison of KLE- and AE-based subspace's diversity for varying shape descriptors, and (d) percentage of invalid designs present in KLE- and AE-based subspaces for varying shape descriptors.}
\label{aerofoil_KLE_AE}
\end{figure*}

\begin{figure*}[hbt!]
\centering
\includegraphics[width=01\textwidth]{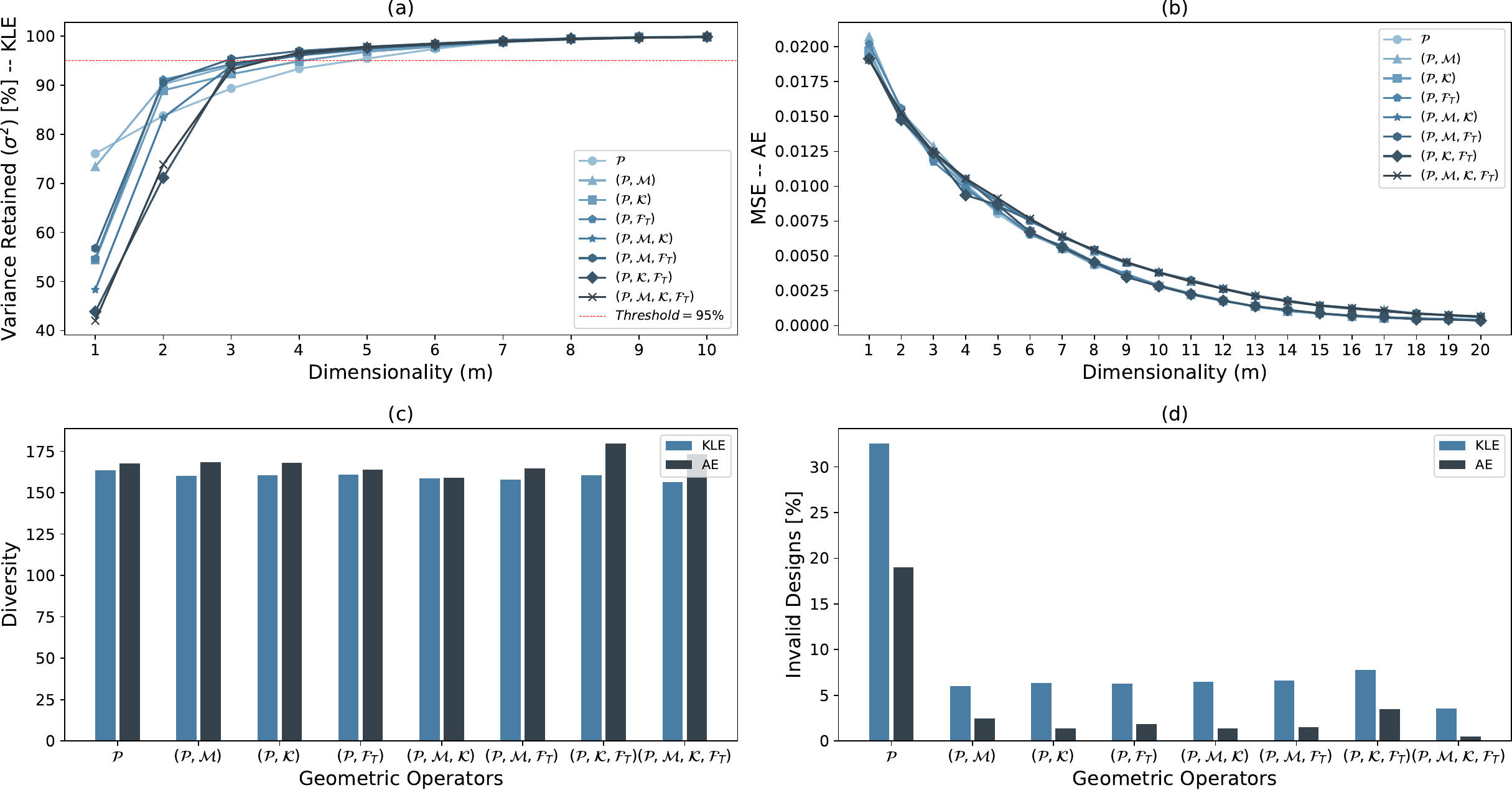}
\caption{Ship Hull Design Spaces: (a) retained percentage of variance with respect to KLE-based subspace's dimension; the horizontal red line indicates the 95\% threshold. (b) Reconstruction accuracy with respect to AE-based subspace's dimension measured via MSE. (c) Comparison of KLE- and AE-based subspace's diversity for varying shape descriptors, and (d) percentage of invalid designs present in KLE- and AE-based subspaces for varying shape descriptors.}
\label{hull_KLE_AE}
\end{figure*}

Figures~\ref{aerofoil_KLE_AE} and \ref{hull_KLE_AE} summarise the obtained results for the latent spaces produced with AEs and the  KLE approaches, for the aerofoil and hull design case, respectively. Latent spaces with varying dimensions produced by varying shape descriptors, i.e., GOs combinations are compared. 

\noindent\textbf{DR results for aerofoil test case:} Figure~\ref{aerofoil_KLE_AE}(a) shows the percentage of retained geometric variance ($\sigma^2$) over the possible latent space dimensions resulting from KLE. Following the recommendation in~\cite{diez2015design}, we select the subspace that retains at least 95\% of variance. Hence, for the majority of shape descriptors, the original 11-dimensional space can be represented by a 4-dimensional latent space while retaining the threshold variance, i.e., $\sigma^2\geq95\%$. The shape descriptor comprising only $\mathcal{P}$, which requires 5 dimensions, and the  $(\mathcal{P},\mathcal{K},\mathcal{F}_T)$ combination, requiring 3 dimensions, are the only exceptions. This results approximately in a 55\% reduction for $(\mathcal{P})$, a 73\% reduction for $(\mathcal{P},\mathcal{K},\mathcal{F}_T)$, and finally, a reduction of 64\% for the remaining GOs combinations. Hence, the introduction of additional GOs in the shape descriptor can lead to a reduction of 1-2 dimensions of the latent space when compared to the classical approach.

In AEs, the selection of subspace dimensionality is performed in a different manner; typically by observing the dataset reconstruction accuracy with increasing dimensions. The obtained results are depicted in Figure~\ref{aerofoil_KLE_AE}(b). The inclusion of $\mathcal{M}$, $\mathcal{F}_T$, as well as the combination of all proposed GOs leads to lower MSE values for the same latent space dimension, when compared to the baseline case where only $\mathcal{P}$ is used. Once again the difference between the standard and enhanced latent spaces is one dimension less, similar to the KLE case mentioned above, e.g., 8 vs 9 or 9 vs 10 depending on the MSE threshold set. One may recall here that for AEs the coordinates of the profile's discretisation are used in $\mathcal{P}$ and therefore the dimensionality reduction is very high. 

Finally, the diversity and validity of the selected design spaces from the KLE- and AE-based approaches are depicted in Figures~\ref{aerofoil_KLE_AE}(c) and (d), respectively. While diversity seems to be almost constant across different shape descriptors, the KLE-based subspaces are clearly superior to the ones produced by AEs. More importantly, invalid designs are practically eliminated when moments or total curvature are introduced to the shape descriptors used in KLE, while this is hardly the case of AEs. Specifically, the inclusion of any of the proposed GOs has a positive impact on the reduction of invalid designs, when compared to the reference case, but in the AE-produced subspaces the percentage of invalid designs remains relatively high, with the only exception being the full set of GOs that bring it down to approximately 0.5\%. One may also observe total energy, $\mathcal{F}_T$, alone does not bring any significant improvements to either KLE- or AE-produced subspaces which is in alignment with the results of the surrogate models training in Figure~\ref{aerofoil_train} and could be partially explained by the focus given by Fourier transforms on the boundary representation. 

    
\noindent\textbf{DR results for ship hull test case:} In this case, the KLE-based approach (see Figure~\ref{hull_KLE_AE}(a)) achieves a significant dimensionality reduction, going from 21 down to 4 for most shape descriptors with the exception of $(\mathcal{P})$ and $(\mathcal{P},\mathcal{F}_T)$ that require a slightly higher number, i.e., 5 dimensions for the same variance threshold of $\sigma^2 \geq 95\%$. However in AEs, as can be seen in Figure~\ref{hull_KLE_AE}(b), the required number of subspace dimensions rises to approximately 20 when a sufficiently small MSE value is applied; keep however in mind that as in the aerofoil case, the geometry in AEs is represented via a high-dimensional discretisation. Although some differences can be observed for low dimensional subspaces, the resulting dimensionality reduction is the same, for all shape descriptors, when an appropriate MSE threshold is used. Similarly, as can be seen in Figure~\ref{hull_KLE_AE}(c), the diversity of the designs in all subspaces and for both approaches remains practically at the same level with only a slight advantage when AE with a full set GOs shape descriptors is used. The major benefit of including GOs in this design case is seen in the reduction of invalid designs; see Figure \ref{hull_KLE_AE}(d). For both approaches, KLE and AE, the reduction in the number of invalid designs is significant with GOs, with the maximum reduction being achieved with $\left(\mathcal{P},\mathcal{M},\mathcal{K},\mathcal{F}_T\right)$. However, in contrast to the aerofoil case, the AE-produced subspaces are on average of better quality, both in terms of diversity and validity; nevertheless, the KLE subspaces are mostly 4-dimensional, whereas the AE-based subspaces require 20 dimensions which is a significant difference when optimisation is to be considered.

In summary, both ML and non-ML DR approaches benefit from the introduction of augmented shape descriptors, in terms of dimensionality reduction and enhancements in diversity and validity. These benefits are clearly observed in the latent space's ability to provide a valid set of designs for both design cases and approaches. Augmented shape descriptors with the proposed GOs provide a better shape representation that enables the extraction of a compact and enhanced latent representation. These effects are obviously beneficial when optimisation is considered and significant cost-savings can be achieved without sacrificing the quality of the expected optimal design in the latent spaces.

\subsubsection{DR with feature selection}
In the context of dimensionality reduction, the previously described subspace learning approach is computationally less intensive as it does not require performance labels. However, feature selection, through sensitivity analysis (SA) is a generally more informed approach as it can reduce uncertainty and provide a ranking of design features with respect to their impact on the quantity of interest, along with dimensionality reduction; see~\cite{khan2021regional}. However, in many engineering problems, its implementation can be computationally demanding, especially when analytical solutions or cheap surrogate approaches are unavailable and running costly physical simulations becomes mandatory.
    
This hindrance can be partially lifted with the introduction of the proposed GOs, as it will be demonstrated by the corresponding results in this subsection. Due to the physics-informed nature of the proposed GOs, they can fully substitute the actual physics-based performance indicators. Note that in this case, we do not combine GOs with $\mathcal{P}$ as in SM and subspace learning, instead, GOs are used as a substitute for the performance-based labels in the design dataset. GOs can harness the geometric variation of designs in the examined design spaces and to a substantial extent can serve as a prior estimation of parametric sensitivities. This approach significantly reduces computational costs as sensitivities, which are typically learnt directly with physical quantity of interest (QoI), can be replaced by the GO-related sensitivities

In our analysis, we used Sobol analysis~\cite{gmdsa_r32}, which is a global variance-based approach. This method investigates how much of the overall variance of the QoI is achieved due to the variability of a collection of design parameters. This variance is usually measured with first-order indices (a.k.a. main effects) or total-order indices (a.k.a. total effects). The former quantifies the direct contribution to QoI variance from an individual parameter over the entire design space, with the latter approximating the overall contribution of a parameter considering both its direct effect and interactions with all other design parameters. These analyses are performed when QoI is a scalar quantity. However, in the general case, the proposed GOs constitute a vector quantity and therefore covariance decomposition~\cite{gmdsa_r29}, which provides generalised sensitivity indices, is used instead. 

Obviously, when sensitivity indices are available, the subset of highly sensitive parameters can be used as the design vector in shape optimisation, with the remaining parameters kept fixed, thereby accelerating the optimisation process. The selection of sensitive parameters is commonly made on the basis of the threshold value, $\epsilon$, which for complex analyses assumes a value of 0.1 or 0.05 as is described in~\cite{gmdsa_r32}. Any parameter with a sensitivity index value greater or equal to $\epsilon$ is therefore included, and others are kept constant to reduce the problem's dimensionality.

\subsubsection{Feature selection results}
\begin{figure*}[hbt!]
    \centering
    \includegraphics[width=01\textwidth]{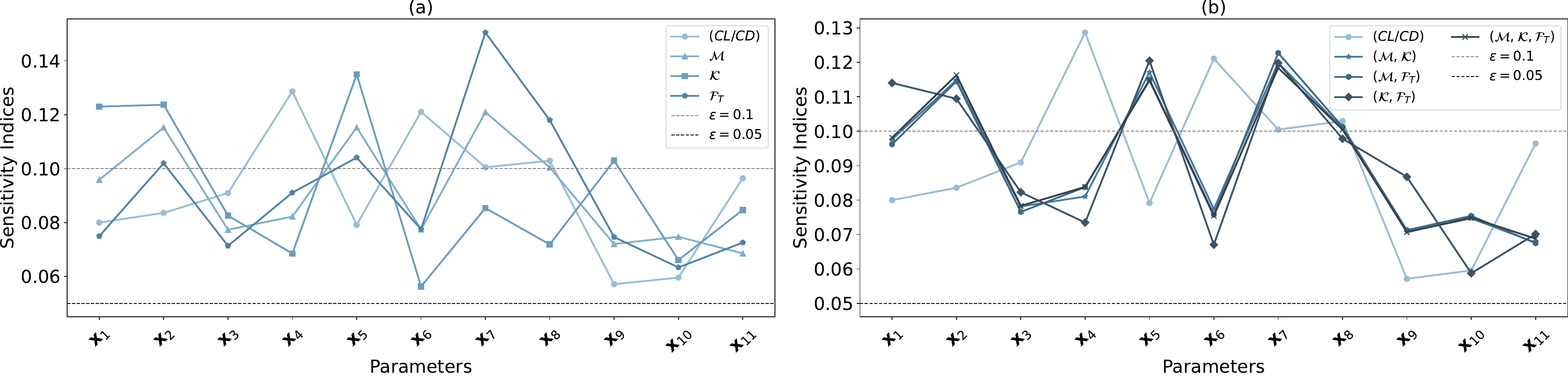}
    \caption{Sensitivity indices of the aerofoil’s 11 design parameters obtained with respect to $C_L/C_D$, $\mathcal{M}$, $\mathcal{K}$, $\mathcal{F}_T$, $\left(\mathcal{M}, \mathcal{K}\right)$, $\left(\mathcal{M}, \mathcal{F}_T\right)$, $\left(\mathcal{K}, \mathcal{F}_T\right)$, and $\left(\mathcal{M}, \mathcal{K}, \mathcal{F}_T\right)$.}
    \label{op_4}
\end{figure*}

Figure~\ref{op_4} compares sensitivity indices for the aerofoil design case with the analysis being perform against $C_L/C_D$ and GOs. To improve figure's clarity, we divided the results into two sub-figures with the left sub-figure, Figure~\ref{op_4}(a), including sensitivity indices for $C_L/C_D$ along with the ones obtained from individual GOs, i.e, $\mathcal{M}$, $\mathcal{K}$, and $\mathcal{F}_T$, while the right sub-figure, Figure~\ref{op_4}(b), compares, once again, the reference indices for $C_L/C_D$ with the ones computed for GOs combinations. In both sub-figures, the $\epsilon=0.1$ and $\epsilon=0.05$ thresholds are marked with dotted lines. Let's begin our discussion by examining the sensitivity results for $C_L/C_D$. At the threshold $\epsilon=0.05$, all parameters are deemed sensitive, indicating they are equally important in influencing the performance of $C_L/C_D$ without necessitating the exclusion of any parameters. Therefore, one may consider adopting a stricter threshold, $\epsilon=0.1$, which shows that out of 11 aerofoil parameters, parameters $\mathbf{x}_4, \mathbf{x}_6, \mathbf{x}_7$, and $\mathbf{x}_8$ are sensitive, while the others can be kept fixed preliminarily to expedite the design process.

\begin{figure*}[hbt!]
    \centering
    \includegraphics[width=0.7\textwidth]{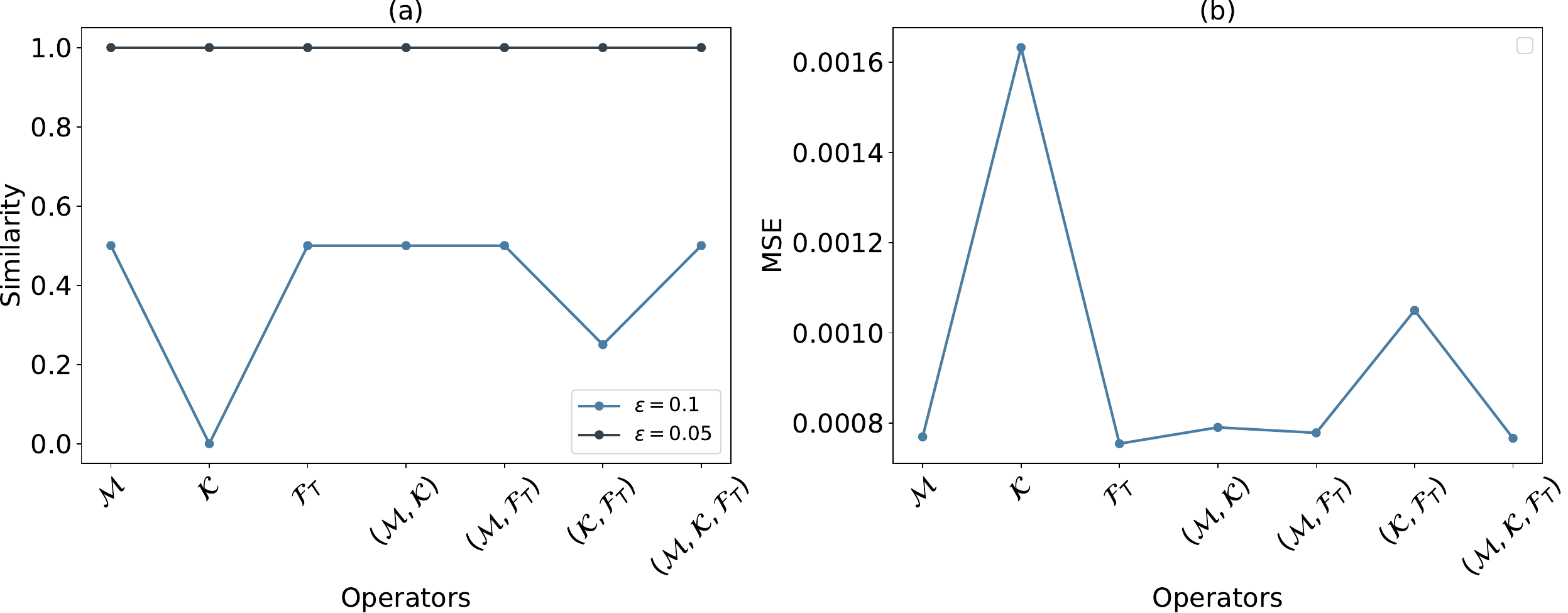}
    \caption{Plot showing (a) similarity and (b) MSE between sensitivity indices obtained using $C_L/C_D$ and GOs: $\mathcal{M}$, $\mathcal{K}$, $\mathcal{F}_T$, $\left(\mathcal{M}, \mathcal{K}\right)$, $\left(\mathcal{M}, \mathcal{F}_T\right)$, $\left(\mathcal{K}, \mathcal{F}_T\right)$, and $\left(\mathcal{M}, \mathcal{K}, \mathcal{F}_T\right)$.}
    \label{op_5}
\end{figure*}

To systematically analyse the correlation between SA results of $C_L/C_D$ and GOs, we first generate a vector with 11 binary variables which correspond to the design parameters used for aerofoil modelling. We then assign a value of 1 to each parameter that achieves a sensitivity index exceeding the $\epsilon$ threshold, and 0 otherwise. Subsequently, for each GO-generated vector, we measure its similarity to the one produced by $C_L/C_D$ indices with cosine similarity. The results of this comparison along with the MSE measured with indices itself are shown in Figure~\ref{op_5}(a) and (b), respectively. The aim of this analysis is to determine the extent of coincidence between parameters that are sensitive to $C_L/C_D$ and the ones that are sensitive to each of the employed GOs. 

In Figure~\ref{op_5}(a) one may easily observe that for the $\epsilon=0.05$ threshold value, all parameters sensitive to $C_L/C_D$ are also sensitive to the employed GOs. However, at $\epsilon=0.1$, the similarity measure drops to 50\% with the exception of $\mathcal{K}$ and $\left(\mathcal{K}, \mathcal{F}_T\right)$, which exhibit an insignificant or lower correlation. The same picture is drawn from Figure~\ref{op_5}(b) when MSE is examined.  Moreover, unlike the results of SMs and DRMs, the individual combination of $\mathcal{F}_T$ provides a good correlation of sensitivities with $C_L/C_D$.
    
\begin{figure*}[hbt!]
        \centering
    \includegraphics[width=01\textwidth]{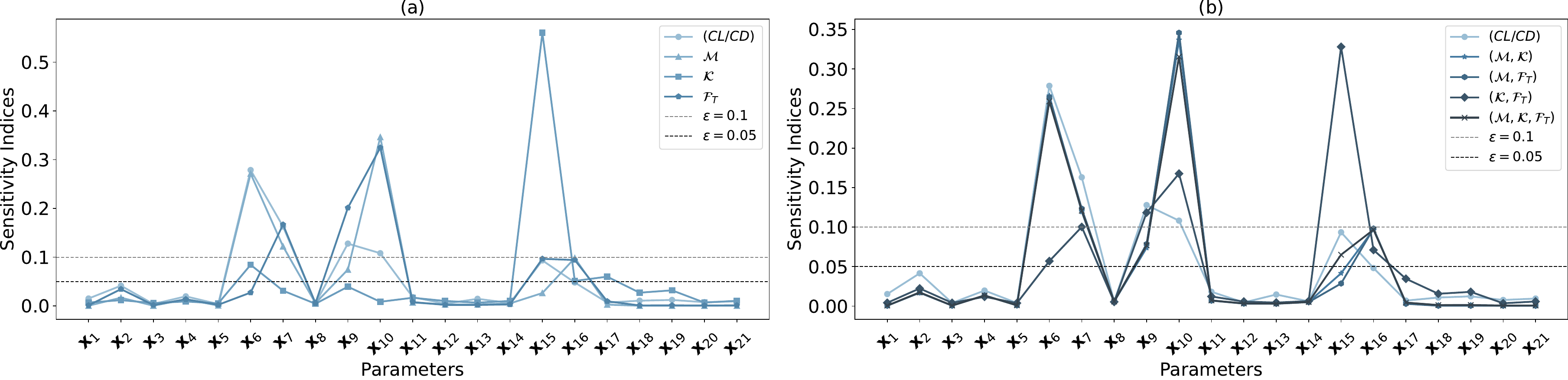}
        \caption{Sensitivity indices of the hull’s 21 design parameters obtained with respect to $C_w$, $\mathcal{M}$, $\mathcal{K}$, $\mathcal{F}_T$, $\left(\mathcal{M}, \mathcal{K}\right)$, $\left(\mathcal{M}, \mathcal{F}_T\right)$, $\left(\mathcal{K}, \mathcal{F}_T\right)$, and $\left(\mathcal{M}, \mathcal{K}, \mathcal{F}_T\right)$.}
        \label{op_1}
    \end{figure*}

\begin{figure*}[hbt!]
        \centering
    \includegraphics[width=0.7\textwidth]{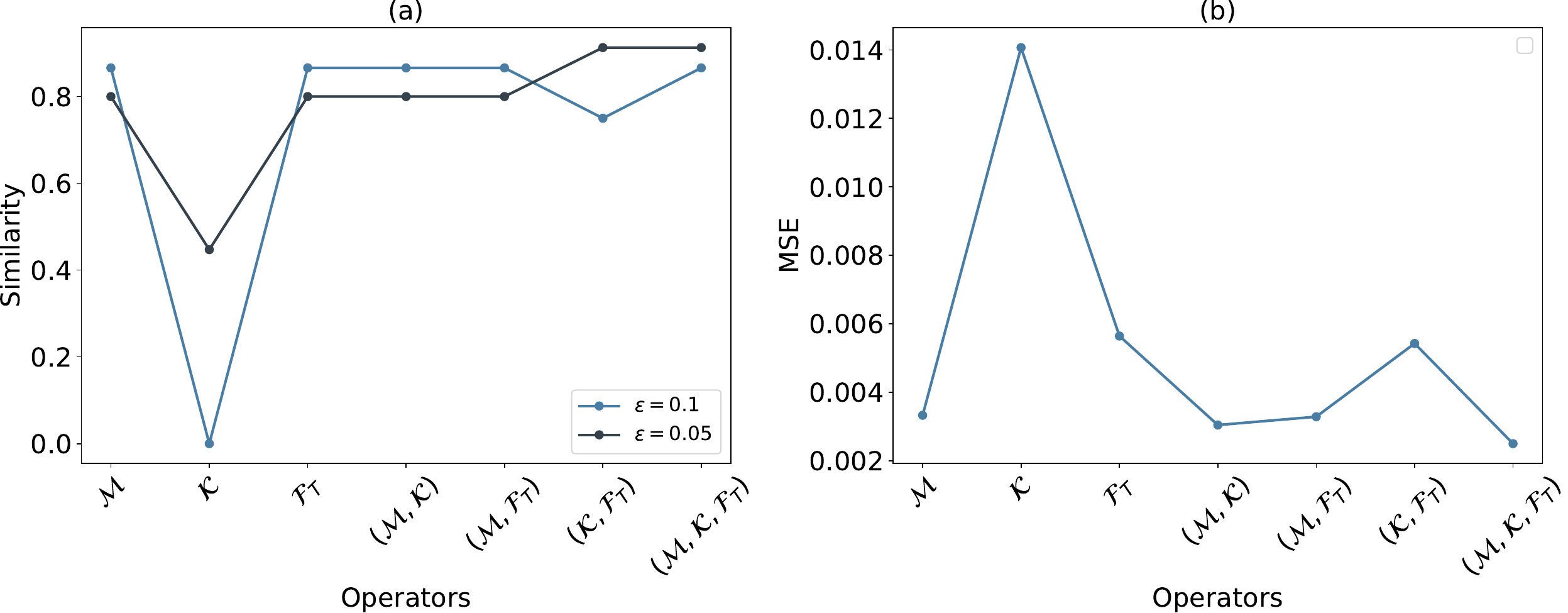}
        \caption{Plot showing (a) similarity and (b) MSE between the sensitivity indices obtained with respect to $C_w$ and GOs: $\mathcal{M}$, $\mathcal{K}$, $\mathcal{F}_T$, $\left(\mathcal{M}, \mathcal{K}\right)$, $\left(\mathcal{M}, \mathcal{F}_T\right)$, $\left(\mathcal{K}, \mathcal{F}_T\right)$, and $\left(\mathcal{M}, \mathcal{K}, \mathcal{F}_T\right)$.}
        \label{op_6}
    \end{figure*}

For the hull design case, the reference sensitivity indices pertain to the wave resistance coefficient $C_w$, and the comparison with GOs is depicted in Figure~\ref{op_1} using a similar arrangement with the one used for the aerofoil designs in Figure~\ref{op_4}. Using the same similarity measure with the aerofoil design case, the similarity and MSE values, with respect to $C_w$, are plotted in Figure \ref{op_6}(a) and (b), respectively. Starting from Figure~\ref{op_6}(a), we can see that the majority of GOs and their combinations have sensitivities that are well aligned with the ones produced for $C_w$. Specifically, for $\epsilon=0.1$ and $\epsilon=0.05$, we get similarities ranging from 87\% and 91\% with one exception, $\mathcal{K}$, that, similar to the aerofoil case, fails to register sufficient similarity. Figure~\ref{op_6}(b) shows the MSE between the sensitivity indices of $C_w$ and GOs, which confirms that utilising all GOs collectively enables a thorough representation of the shape's inherent variability with respect to its geometry and physics, thereby providing a strong correlation with the sensitivity indices of $C_w$.

    In summary, we can highlight the following two important results: First, performing SA with GOs is generally computationally inexpensive, and second, all performance-sensitive parameters remain equally sensitive to the majority of the employed GOs in both design cases. Especially when the full GOs vector is used, the similarity/correlation is guaranteed since the few non-correlated operators do not affect the final result. Thus, performing SA with GOs can give designers a very good and cost-effective estimation of the expected results before proceeding to detailed and expensive actual performance-based analyses.

    \subsection{Generative model training}
    As previously discussed in \S\ref{drawback}, unlike DRMs that solely focus on reducing the dimensions of the parametric design space, GMs employ latent feature extraction for modelling the underlying distribution of the provided design datasets. Therefore, they can be seen as data-driven parameterisation approaches where the latent variables are used as parameters for forming generative design spaces. These spaces are not only low dimensional, which obviously expedites shape optimisation, but, if properly trained, can also produce novel and valid designs beyond the range of samples residing in the training dataset.

    However, most current state-of-the-art GMs lack a mechanism for incorporating a notion of physics/performance indices against which designs are optimised. Furthermore, since both the input and output of these models consist of low-level 3D shape representations, they fail to capture structural and shape characteristics essential for performance analysis. This often results in a lack of surface smoothness and a high number of invalid designs. Consequently, substantial effort may be spent training a generative model only to discover that many of the produced designs are infeasible or fail to meet design requirements.

    \textbf{PaDGAN:} To address these issues, Chen and  Ahmed~\cite{chen2021padgan} proposed an extension of the generative adversarial network (GAN) called PaDGAN. This model introduces a performance-augmented determinantal point process (DPP) loss function which promotes diversity and quality by assigning a lower loss to sets of designs that exhibit both high quality and diversity. Specifically, the PaDGAN model constructs a kernel matrix $L_B$ for a design batch $B$. This matrix is defined as $L_B(i, j) = k(\mathbf{x}_i, \mathbf{x}_j)(q(\mathbf{x}_i)q(\mathbf{x}_j))^{\gamma_0}$, where $\mathbf{x}_i, \mathbf{x}_j \in B$; $q(\mathbf{x})$ represents the quality of $\mathbf{x}$; $k(\mathbf{x}_i, \mathbf{x}_j)$ is the similarity kernel between $\mathbf{x}_i$ and $\mathbf{x}_j$; and $\gamma_0$ is a parameter controlling the weight placed on quality. PaDGAN's training involves the minimisation of the loss function, which is a combination of a baseline loss function used in standard GANs and the DPP loss functions. This augmented loss function is expressed as follows.
    \begin{equation}\label{GANlossFun}
    \min_{G} \max_{D} \left[ \left( \mathbb{E}_{x \sim P_{\text{data}}} [\log D(x)] + \mathbb{E}_{z \sim P_{z}} [\log (1 - D(G(z)))] \right) + \gamma_1 \left( -\frac{1}{|B|} \sum_{i=1}^{|B|} \log \lambda_i \right) \right],
    \end{equation}
    where, $\lambda_i$ is the $i$th eigenvalue of $L_B$, and $\gamma_1$ is a parameter influencing the impact of the DPP loss function. The first part of the equation pertains to the loss of a baseline GAN, comprising two neural networks: a generator $G$ and a discriminator $D$. These networks are trained simultaneously to enable $G$ to effectively map from a chosen latent distribution, $P_{\mathbf{z}}$, to the distribution of the training data, $P_{\text{data}}(\mathbf{x})$. For more detailed information, the interested reader may refer to \cite{chen2021padgan}.

    \textbf{Training PaDGAN with GOs:} In the case of aerofoils, the quality metric $q(\mathbf{x})$ in $L_B$ represents $C_L/C_D$, which must be evaluated during the training of PaDGAN. Direct evaluation of $C_L/C_D$ can impose a significant computational burden, even when using low-fidelity potential flow solvers. Therefore, the authors in \cite{chen2021padgan} trained a surrogate model during the offline stage and integrated it with PaDGAN during the training to evaluate the gradients of $C_L/C_D$ and compute the loss function in Eq. \eqref{GANlossFun}. However, this study demonstrates that, due to the physics-informed nature of the proposed GOs, one could use them as a substitute for $C_L/C_D$ and still achieve similar performance compared to the baseline PaDGAN. The version of PaDGAN trained with $(\mathcal{M}, \mathcal{K}, \mathcal{F}_T)$ is referred to as PaDGAN-GO. It is worth noting that in our case, GOs are a vector quantity whereas $C_L/C_D$ is scalar; therefore, to calculate $L_B$, we take the norm of $(\mathcal{M}, \mathcal{K}, \mathcal{F}_T)$, i.e., $q(\mathbf{x}) = \|(\mathcal{M}, \mathcal{K}, \mathcal{F}_T)\|_1$, as recommended by \cite{chen2021padgan}.

    We compare the performance of PaDGAN-GO and PaDGAN, following the same methodology used in \cite{chen2021padgan} which was used to compare PaDGAN with GAN. The comparison is based on diversity, quality, and novelty scores of the designs resulting from the trained models. The diversity score is expressed as the mean log determinant of the similarity matrix $k(\mathbf{x}_i, \mathbf{x}_j)$, while quality is measured with the average $C_L/C_D$ value of the generated designs. Finally, the novelty score, is measured by the distance of each generated design to its nearest training sample. For more details on these metrics, see~\cite{chen2021padgan}. Note that for PaDGAN-GO, we used $q(\mathbf{x}) = \|(\mathcal{M}, \mathcal{K}, \mathcal{F}_T)\|_1$ during training; however, for measuring the quality score, we calculated the actual $C_L/C_D$ values for a fair comparison with PaDGAN. Additionally, similar to previous model cases, we also  measure the percentage of invalid designs resulting from these models.

\begin{figure*}[hbt!]
    \centering
\includegraphics[width=01\textwidth]{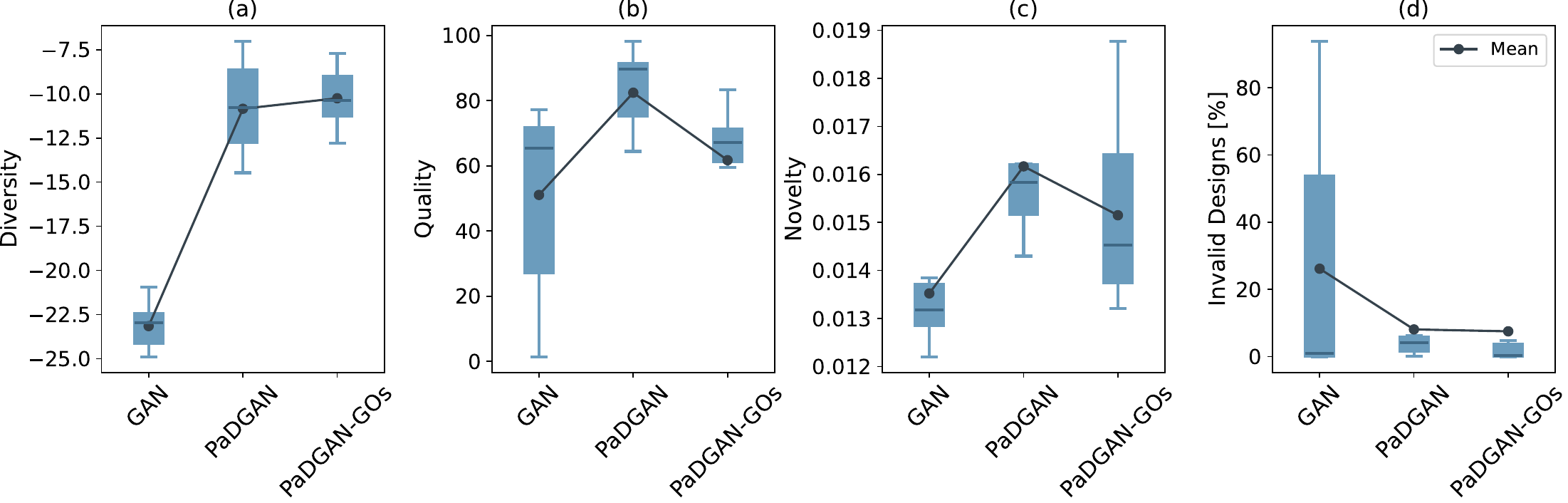}
    \caption{(a) Diversity, (b) quality, (c) novelty scores and (d) percentage of invalid designs resulting from the trained GAN, PaDGAN and PaDGAN-GO models.}
    \label{op_25}
\end{figure*}

\textbf{Comparison between GAN, PaDGAN and PaDGAN-GO:} Training for all three models was performed with the same training dataset and under the same settings as described in \cite{chen2021padgan}. The resulting scores for GAN, PaDGAN, and PaDGAN-GO are shown in Figure~\ref{op_25} which illustrates that across all employed metrics, GAN has clearly the worst performance. Although there are no significant differences between PaDGAN and PaDGAN-GO, PaDGAN slightly outperforms PaDGAN-GO in quality and novelty. However, it is important to remember that PaDGAN-GO is trained with GOs, which are computationally less expensive to evaluate compared to directly evaluating physics, specifically $C_L/C_D$, in this case. This illustrates that PaDGAN-GO provides high-quality designs, which are on par with PaDGAN and once again confirms the physics-informed nature of the proposed GOs. 
    
\section{Summary of results and concluding remarks}
    In this work, we propose the use of physics-informed GOs constructed upon the shape’s differential and/or integral properties retrieved via Fourier analysis, curvature, and geometric moments to enrich the geometric data provided to the employ different model setups including SMs, DRMs, and GMs in engineering applications. 
    
    Through extensive experimentation, we show that GOs enable the extraction of useful high-level shape characteristics, even when using simple model architectures or low-level data representations like design parameters.  These operators capture both global and local shape features by explicitly encoding the relevant shape information. This not only augments the training dataset with a compact geometric representation of free-form shapes but also embeds physical information. 

    During the surrogate modelling via SMs, the inclusion of GOs highlights the performance of neural networks NNs and GPRs in modelling the lift-to-drag ratio ($C_L/C_D$) of aerofoils and wave-resistance ($C_w$) of ship hulls. The results demonstrate that using the design parameter vector $\mathcal{P}$ alone yields low $R^2$ values for both NNs and GPRs, indicating that $\mathcal{P}$ is insufficient for capturing the complexities of $C_L/C_D$ and $C_w$. The incorporation of $\mathcal{M}$ into the feature set ($\left(\mathcal{P},\mathcal{M}\right)$) significantly enhances model performance, as evidenced by substantial increases in $R^2$ and decreases in MAPE and RMSE. This suggests that $\mathcal{M}$ is a critical predictive feature for accurately modelling $C_L/C_D$ and $C_w$.

    Further, the addition of $\mathcal{K}$ and $\mathcal{F}_T$, either individually or jointly ($\left(\mathcal{P},\mathcal{K},\mathcal{F}_T\right)$), does not improve model accuracy, thereby indicating them as non-contributive features. In both test cases, the pattern of error metrics for NNs is consistent with that observed in the aerofoil case, achieving the highest improvement with $(\mathcal{P},\mathcal{M},\mathcal{F}_T)$.
 This configuration suggests that $\mathcal{M}$ plays a pivotal role in capturing the nonlinear dynamics of $C_L/C_D$ and $C_w$.

In case of dimension reduction with subspace learning, the training of KLE and AE with and without GOs are evaluated based on the resulting subspace's ability to produce valid and diverse designs across different subspace dimensionalities. Results showed that KLE effectively retains a significant percentage of variance ($\sigma^2 \geq 95\%$) at lower dimensionalities when GOs are integrated.
Without GOs, using only $\mathcal{P}$, a 55\% reduction in the original space dimensionality is achieved, while the introduction of GOs generally enhances this reduction, peaking at 73\% for $(\mathcal{P},\mathcal{K},\mathcal{F}_T)$.
The optimal subspace dimensionalities for various GOs combinations range between 3 and 4 dimensions, significantly reducing from the original 11 dimensions.

The AE's reconstruction accuracy generally improves with the inclusion of GOs, evidenced by lower MSE in comparison to the baseline $\mathcal{P}$ only scenario.
The reduced dimensionality for optimal AE performance with the inclusion of all GOs is one less than without GOs, suggesting a better representation with a slightly reduced subspace. 
Both KLE and AE, when integrated with GOs, particularly $\left(\mathcal{P},\mathcal{M}, \mathcal{K}, \mathcal{F}_T\right)$, significantly reduced the occurrence of invalid designs compared to the baseline case that used only geometry, i.e., $\mathcal{P}$. The diversity of designs remains constant across both methods, but KLE generally offers better subspace quality in terms of validity and dimensionality.

DR with feature selection via sensitivity analysis revealed that for the aerofoil, all parameters sensitive to $C_L/C_D$ were also sensitive to GOs at the stricter threshold of $\epsilon = 0.05$. However, at the looser threshold of $\epsilon = 0.1$, the sensitivity correlation dropped, particularly for $\mathcal{K}$ and its combinations, indicating less consistency. Notably, the combination of all GOs showed the lowest mean squared error (MSE), suggesting that a comprehensive approach to GOs is more reflective of actual sensitivity. Conversely, $\mathcal{K}$ individually showed the least similarity and consequently the highest MSE, implying significant variability in its impact on aerodynamic performance.

The hull test case displayed a high degree of similarity between the sensitivity indices for $C_w$ and the combined GOs, with 87\% and 91\% similarity at the thresholds of $\epsilon = 0.1$ and $0.05$, respectively. This indicates that GOs are particularly effective in capturing the performance sensitivities related to hull geometry and physics, as shown by the consistently low MSE when all GOs are used together. However, similar to the aerofoil case, $\mathcal{K}$ and its combinations again stood out for their divergence from the general sensitivity pattern.

Finally, to demonstrate the added benefits of GOs in GMs, we selected a PaDGAN model whose training involves a hybrid approach using both traditional GAN loss and performance-augmented determinantal point process (DPP) loss. This balances the generation of designs that are not only diverse but also adhere closely to quality metrics. For enhancing diversity and quality, PaDGAN incorporates a physics-based metric, such as $C_L/C_D$ for aerofoils, directly evaluated during training. This can result in extensive training complexity, even when approximated through surrogate models. Therefore, we propose PaDGAN-GO, which transitions to using GOs as a substitute, proving to be computationally more efficient while maintaining performance levels comparable to direct physical evaluations. While traditional GANs performed poorly across these metrics, PaDGAN and PaDGAN-GO showed robust performance with little difference between them, demonstrating that the integration of GOs does not compromise the quality or diversity of designs.

In conclusion, the results suggest that integrating physics-informed operators like GOs in GMs not only reduces computational overhead but also maintains, if not enhances, the capability to generate viable, high-quality designs, underscoring the potential of advanced generative models in design optimisation contexts.
    
    \subsection{Future work}
    For future exploration, we are particularly interested in investigating the potential of integrating geometric operators with neural operators. Typically, standard learning models only learn mappings between inputs and outputs of fixed dimensions. These models lack the expressive power necessary to handle mappings between functions on continuous domains—termed as operators. In contrast, neural operators are not constrained by the grid of training data and can predict outcomes at any point within the output domain. This is achieved by approximating the underlying operator, which represents the mapping between input and output function spaces, each potentially infinite-dimensional. By integrating neural operators with our geometric operators, we aim to enhance their ability to incorporate shape-informed characteristics, which are crucial for applications in shape optimisation. Furthermore, we are also interested in exploring the integration of geometric operators to improve the learning of neural implicit representations of free-form shapes. 
    
    \section*{Acknowledgements}
    This work received funding from: 
    \begin{enumerate}
    \item European Union's Horizon 2020 research and innovation programme under the Marie Skłodowska-Curie grant agreement No 860843 – \say{GRAPES: Learning, Processing and Optimising Shapes}, PI for the University of Strathclyde: P.D. Kaklis, and
    \item Nazarbayev University, Kazakhstan, under the Faculty Development Competitive Research Grants Program 2022–2024: “Shape Optimization of Free-form Functional surfaces using isogeometric Analysis and Physics-Informed Surrogate Models—SOFFA-PHYS”, Grant Award No. 11022021FD2927, PI: K.V. Kostas
    \end{enumerate}    

    \bibliographystyle{elsarticle-num}
    \bibliography{01ref}

\begin{thebibliography}{10}
\expandafter\ifx\csname url\endcsname\relax
  \def\url#1{\texttt{#1}}\fi
\expandafter\ifx\csname urlprefix\endcsname\relax\def\urlprefix{URL }\fi
\expandafter\ifx\csname href\endcsname\relax
  \def\href#1#2{#2} \def\path#1{#1}\fi

\bibitem{shukla2024deep}
K.~Shukla, V.~Oommen, A.~Peyvan, M.~Penwarden, N.~Plewacki, L.~Bravo, A.~Ghoshal, R.~M. Kirby, G.~E. Karniadakis, Deep neural operators as accurate surrogates for shape optimization, Engineering Applications of Artificial Intelligence 129 (2024) 107615.

\bibitem{poole2017high}
D.~Poole, C.~Allen, T.~Rendall, High-fidelity aerodynamic shape optimization using efficient orthogonal modal design variables with a constrained global optimizer, Computers \& Fluids 143 (2017) 1--15.

\bibitem{karniadakis2021physics}
G.~E. Karniadakis, I.~G. Kevrekidis, L.~Lu, P.~Perdikaris, S.~Wang, L.~Yang, Physics-informed machine learning, Nature Reviews Physics 3~(6) (2021) 422--440.

\bibitem{nikishova2020sensitivity}
A.~Nikishova, G.~E. Comi, A.~G. Hoekstra, Sensitivity analysis based dimension reduction of multiscale models, Mathematics and Computers in Simulation 170 (2020) 205--220.

\bibitem{diez2015design}
M.~Diez, E.~F. Campana, F.~Stern, {Design-space dimensionality reduction in shape optimization by Karhunen--Lo{\`e}ve expansion}, Computer Methods in Applied Mechanics and Engineering 283 (2015) 1525--1544.

\bibitem{goodfellow2016deep}
I.~Goodfellow, Y.~Bengio, A.~Courville, Deep Learning, Adaptive Computation and Machine Learning series, MIT Press, 2016.

\bibitem{weisz2023toward}
J.~D. Weisz, M.~Muller, J.~He, S.~Houde, Toward general design principles for generative {AI} applications, in: Joint Workshops on Human-AI Co-Creation with Generative Models and User-Aware Conversational Agents, 2023, pp.~--.

\bibitem{gordon2020combining}
J.~Gordon, J.~M. Hern{\'a}ndez-Lobato, Combining deep generative and discriminative models for bayesian semi-supervised learning, Pattern Recognition 100 (2020) 107156.

\bibitem{ryan1997modern}
T.~Ryan, Modern Regression Methods, A Wiley-Interscience publication, Wiley, 1997.

\bibitem{KHAN2022103327}
S.~Khan, P.~Kaklis, A.~Serani, M.~Diez, K.~Kostas, Shape-supervised dimension reduction: Extracting geometry and physics associated features with geometric moments, Computer-Aided Design 150 (2022) 103327.

\bibitem{10.3390/jmse11101851}
Z.~Masood, K.~Kostas, S.~Khan, P.~Kaklis, Shape-informed dimensional reduction in airfoil/hydrofoil modeling, J. Mar. Sci. Eng. 11~(1851) (2023).

\bibitem{tomczak2022deep}
J.~Tomczak, Deep Generative Modeling, Springer International Publishing, 2022.

\bibitem{khan2023shiphullgan}
S.~Khan, K.~Goucher-Lambert, K.~Kostas, P.~Kaklis, Shiphullgan: A generic parametric modeller for ship hull design using deep convolutional generative model, Computer Methods in Applied Mechanics and Engineering 411 (2023) 116051.

\bibitem{chen2021padgan}
W.~Chen, F.~Ahmed, Padgan: Learning to generate high-quality novel designs, Journal of Mechanical Design 143~(3) (2021) 031703.

\bibitem{iliadis2023engineering}
L.~Iliadis, I.~Maglogiannis, S.~Alonso, C.~Jayne, E.~Pimenidis, Engineering Applications of Neural Networks: 24th International Conference, EAAAI/EANN 2023, Le{\'o}n, Spain, June 14--17, 2023, Proceedings, Communications in Computer and Information Science, Springer Nature Switzerland, 2023.

\bibitem{regenwetter2022deep}
L.~Regenwetter, A.~H. Nobari, F.~Ahmed, Deep generative models in engineering design: A review, Journal of Mechanical Design 144~(7) (2022) 071704.

\bibitem{rubinstein1997discriminative}
Y.~D. Rubinstein, T.~Hastie, et~al., Discriminative vs informative learning., in: KDD-97 (American Association for Artificial Intelligence), Vol.~5, 1997, pp. 49--53.

\bibitem{granadeiro2013general}
V.~Granadeiro, L.~Pina, J.~P. Duarte, J.~R. Correia, V.~M. Leal, A general indirect representation for optimization of generative design systems by genetic algorithms: Application to a shape grammar-based design system, Automation in Construction 35 (2013) 374--382.

\bibitem{sindhu2020survey}
K.~Sindhu~Meena, S.~Suriya, A survey on supervised and unsupervised learning techniques, in: Proceedings of international conference on artificial intelligence, smart grid and smart city applications: AISGSC 2019, Springer, 2020, pp. 627--644.

\bibitem{reeves1988three}
A.~P. Reeves, R.~J. Prokop, S.~E. Andrews, F.~P. Kuhl, {Three-dimensional shape analysis using moments and Fourier descriptors}, IEEE Transactions on Pattern Analysis and Machine Intelligence 10~(6) (1988) 937--943.

\bibitem{joseph2019momen}
M.~Joseph-Rivlin, A.~Zvirin, R.~Kimmel, Momen$^e$t: Flavor the moments in learning to classify shapes, in: Proceedings of the IEEE/CVF international conference on computer vision workshops, 2019, pp. 0--0.

\bibitem{li2020ggm}
D.~Li, X.~Shen, Y.~Yu, H.~Guan, H.~Wang, D.~Li, {GGM-net}: Graph geometric moments convolution neural network for point cloud shape classification, IEEE Access 8 (2020) 124989--124998.

\bibitem{ye2021curvature}
Z.~Ye, N.~Umetani, T.~Igarashi, T.~Hoffmann, A curvature and density-based generative representation of shapes, in: Computer Graphics Forum, Vol.~40, Wiley Online Library, 2021, pp. 38--53.

\bibitem{gmdsa_r5}
A.~Krishnamurthy, S.~McMains, Accurate {GPU}-accelerated surface integrals for moment computation, Computer-Aided Design 43~(10) (2011) 1284--1295.

\bibitem{ssdr_r5}
A.~Taber, G.~Kumar, M.~Freytag, V.~Shapiro, A moment-vector approach to interoperable analysis, Computer-Aided Design 102 (2018) 139--147.

\bibitem{shen2017computational}
X.~Shen, E.~Avital, M.~A. Rezaienia, G.~Paul, T.~Korakianitis, Computational methods for investigation of surface curvature effects on airfoil boundary layer behavior, Journal of Algorithms \& Computational Technology 11~(1) (2017) 68--82.

\bibitem{diaconis1987application}
P.~Diaconis, Application of the method of moments in probability and statistics, Moments in mathematics 37 (1987) 125--142.

\bibitem{ssdr_r34}
D.~F. Atrevi, D.~Vivet, F.~Duculty, B.~Emile, A very simple framework for {3D} human poses estimation using a single {2D} image: Comparison of geometric moments descriptors, Pattern Recognition 71 (2017) 389--401.

\bibitem{gmdsa_r8}
A.~M. Bronstein, M.~M. Bronstein, R.~Kimmel, Numerical geometry of non-rigid shapes, Springer Science \& Business Media, 2008.

\bibitem{teh1988image}
C.-H. Teh, R.~T. Chin, On image analysis by the methods of moments, IEEE Transactions on pattern analysis and machine intelligence 10~(4) (1988) 496--513.

\bibitem{milanfar1995reconstructing}
P.~Milanfar, G.~C. Verghese, W.~C. Karl, A.~S. Willsky, Reconstructing polygons from moments with connections to array processing, IEEE Transactions on signal processing 43~(2) (1995) 432--443.

\bibitem{ssdr_r29}
G.~Kumar, A.~Taber, An integral representation of fields with applications to finite element analysis of spatially varying materials, Computer-Aided Design 126 (2020) 102869.

\bibitem{ssdr_r7}
P.~Jin, B.~Xie, F.~Xiao, Multi-moment finite volume method for incompressible flows on unstructured moving grids and its application to fluid-rigid body interactions, Computers \& Structures 221 (2019) 91--110.

\bibitem{shohat1950problem}
J.~A. Shohat, J.~D. Tamarkin, The problem of moments, Vol.~1, American Mathematical Society (RI), 1950.

\bibitem{gmdsa_r58}
B.~Gustafsson, C.~He, P.~Milanfar, M.~Putinar, Reconstructing planar domains from their moments, Inverse Problems 16~(4) (2000) 1053.

\bibitem{gmdsa_r59}
A.~Kousholt, J.~Schulte, Reconstruction of convex bodies from moments, Discrete \& Computational Geometry 65~(1) (2021) 1--42.

\bibitem{gmdsa_r62}
H.-C. Kim, On the volumetric balanced variation of ship forms, Journal of Ocean Engineering and Technology 27~(2) (2013) 1--7.

\bibitem{gmdsa_r61}
S.~Han, Y.-S. Lee, Y.~B. Choi, Hydrodynamic hull form optimization using parametric models, Journal of marine science and technology 17~(1) (2012) 1--17.

\bibitem{gmdsa_r66}
E.~O. Tuck, Shallow-water flows past slender bodies, Journal of fluid mechanics 26~(1) (1966) 81--95.

\bibitem{gmdsa_r68}
J.~V. Wehausen, The wave resistance of ships, in: Advances in applied mechanics, Vol.~13, Elsevier, 1973, pp. 93--245.

\bibitem{fu2008shape}
J.~Fu, S.~B. Joshi, T.~W. Simpson, Shape differentiation of freeform surfaces using a similarity measure based on an integral of gaussian curvature, Computer-Aided Design 40~(3) (2008) 311--323.

\bibitem{hildebrandt2004anisotropic}
K.~Hildebrandt, K.~Polthier, Anisotropic filtering of non-linear surface features, in: Computer Graphics Forum, Vol.~23, Wiley Online Library, 2004, pp. 391--400.

\bibitem{chen2024fdspc}
Z.~Chen, Y.~Li, Fdspc: Fast and direct smooth path planning via continuous curvature integration, arXiv preprint arXiv:2405.03281 (2024).

\bibitem{nemnem2014smooth}
A.~F. Nemnem, M.~G. Turner, K.~Siddappaji, M.~Galbraith, A smooth curvature-defined meanline section option for a general turbomachinery geometry generator, in: Turbo Expo: Power for Land, Sea, and Air, Vol. 45615, American Society of Mechanical Engineers, 2014, p. V02BT39A026.

\bibitem{korakianitis1992surface}
T.~Korakianitis, P.~Papagiannidis, Surface-curvature-distribution effects on turbine-cascade performance, in: Turbo Expo: Power for Land, Sea, and Air, Vol. 78934, American Society of Mechanical Engineers, 1992, p. V001T01A044.

\bibitem{10.1115/1.4005969}
T.~Korakianitis, M.~A. Rezaienia, I.~A. Hamakhan, E.~J. Avital, J.~J.~R. Williams, {Aerodynamic Improvements of Wind-Turbine Airfoil Geometries With the Prescribed Surface Curvature Distribution Blade Design (CIRCLE) Method }, Journal of Engineering for Gas Turbines and Power 134~(8) (2012) 082601.

\bibitem{massardo1989axial_I}
A.~Massardo, A.~Satta, Axial flow compressor design optimization: {P}art {I}—pitchline analysis and multivariate objective function influence, in: Turbo Expo: Power for Land, Sea, and Air, Vol. 79139, American Society of Mechanical Engineers, 1989, p. V001T01A081.

\bibitem{massardo1989axial_II}
A.~Massardo, A.~Satta, M.~Marini, Axial flow compressor design optimization: {P}art {II}—through-flow analysis, in: Turbo Expo: Power for Land, Sea, and Air, Vol. 79139, American Society of Mechanical Engineers, 1989, p. V001T01A082.

\bibitem{song2014effects}
Y.~Song, C.~Gu, Effects of curvature continuity of compressor blade profiles on their performances, in: Turbo Expo: Power for Land, Sea, and Air, Vol. 45608, American Society of Mechanical Engineers, 2014, p. V02AT37A020.

\bibitem{park1987three}
K.~S. Park, N.~S. Lee, {A three-dimensional Fourier descriptor for human body representation/reconstruction from serial cross sections}, Computers and biomedical research 20~(2) (1987) 125--140.

\bibitem{gmdsa_r21}
K.~Kostas, A.~Ginnis, C.~Politis, P.~Kaklis, Ship-hull shape optimization with a {T}-spline based {BEM}--isogeometric solver, Computer Methods in Applied Mechanics and Engineering 284 (2015) 611--622.

\bibitem{khan2022geometric}
S.~Khan, P.~Kaklis, A.~Serani, M.~Diez, Geometric moment-dependent global sensitivity analysis without simulation data: application to ship hull form optimisation, Computer-Aided Design 151 (2022) 103339.

\bibitem{gmdsa_r22}
K.~Belibassakis, T.~P. Gerostathis, K.~Kostas, C.~Politis, P.~Kaklis, A.~Ginnis, C.~Feurer, A {BEM}-isogeometric method for the ship wave-resistance problem, Ocean Engineering 60 (2013) 53--67.

\bibitem{GINNIS2014425}
A.~Ginnis, K.~Kostas, C.~Politis, P.~Kaklis, K.~Belibassakis, T.~Gerostathis, M.~Scott, T.~Hughes, {Isogeometric boundary-element analysis for the wave-resistance problem using T-splines}, Computer Methods in Applied Mechanics and Engineering 279 (2014) 425--439.

\bibitem{KOSTAS201779}
K.~Kostas, A.~Ginnis, C.~Politis, P.~Kaklis, Shape-optimization of {2D} hydrofoils using an isogeometric bem solver, Computer-Aided Design 82 (2017) 79--87, isogeometric Design and Analysis.

\bibitem{doi:10.2514/1.29958}
B.~M. Kulfan, Universal parametric geometry representation method, Journal of Aircraft 45~(1) (2008) 142--158.

\bibitem{chen2020airfoil}
W.~Chen, K.~Chiu, M.~D. Fuge, Airfoil design parameterization and optimization using b{\'e}zier generative adversarial networks, AIAA journal 58~(11) (2020) 4723--4735.

\bibitem{Drela87}
M.~Drela, M.~Giles, Viscous-inviscid analysis of transonic and low reynolds number airfoils, AIAA 25~(10) (1987) 1347--1355.

\bibitem{Drela89}
M.~Drela, {XFOIL: An Analysis and Design System for Low Reynolds Number Airfoils}, in: T.~Mueller (Ed.), Low Reynolds Number Aerodynamics. Lecture Notes in Engineering, Vol.~54, Springer, Berlin, Heidelberg, 1989, pp. 1--12.

\bibitem{piffl_r23}
C.~K. Williams, C.~E. Rasmussen, Gaussian processes for machine learning, Vol.~2, MIT press Cambridge, MA, 2006.

\bibitem{khan2018sampling}
S.~Khan, E.~Gunpinar, Sampling {CAD} models via an extended teaching--learning-based optimization technique, Computer-Aided Design 100 (2018) 52--67.

\bibitem{gmdsa_r32}
X.-Y. Zhang, M.~N. Trame, L.~J. Lesko, S.~Schmidt, Sobol sensitivity analysis: a tool to guide the development and evaluation of systems pharmacology models, CPT: pharmacometrics \& systems pharmacology 4~(2) (2015) 69--79.

\bibitem{masood2024generative}
Z.~Masood, M.~Usama, S.~Khan, K.~Kostas, P.~D. Kaklis, Generative vs. non-generative models in engineering shape optimization, Journal of Marine Science and Engineering 12~(4) (2024) 566.

\bibitem{khan2021regional}
S.~Khan, P.~Kaklis, From regional sensitivity to intra-sensitivity for parametric analysis of free-form shapes: Application to ship design, Advanced Engineering Informatics 49 (2021) 101314.

\bibitem{gmdsa_r29}
F.~Gamboa, A.~Janon, T.~Klein, A.~Lagnoux, Sensitivity indices for multivariate outputs, Comptes Rendus Mathematique 351~(7) (2013) 307--310.

\end{thebibliography}
    \end{document}